\documentclass[journal]{IEEEtran}
\ifCLASSINFOpdf
\else
\fi

\usepackage{graphicx}
\usepackage{amsmath,amssymb} 
\usepackage{color}
\newcommand{\argmin}{\operatornamewithlimits{argmin}}

\usepackage{algorithm}
\usepackage{algpseudocode}
\usepackage{graphics}
\usepackage{multirow}
\usepackage{epsfig}
\usepackage{url}
\usepackage{lineno}

\begin{document}
%
\title{Fast Landmark Localization \\with 3D Component Reconstruction and CNN for Cross-Pose Recognition}
%
%
%

\author{$^*$Gee-Sern~(Jison)~Hsu, Hung-Cheng Shie, Cheng-Hua Hsieh
\thanks{Authors are with the Artificial Vision Laboratory,}
\thanks{National Taiwan University of Science and Technology}
\thanks{43 Sec.4 Keelung Rd., Daan Dist., Taipei 10607, Taiwan}
\thanks{E-mail: $^*$jison@mail.ntust.edu.tw}
}

%
%

\markboth{IEEE TRANSACTIONS ON CIRCUITS AND SYSTEMS FOR VIDEO TECHNOLOGY}{}
%



\maketitle

\begin{abstract}
Two approaches are proposed for cross-pose face recognition, one is based on the 3D reconstruction of facial components and the other is based on the deep Convolutional Neural Network (CNN). Unlike most 3D approaches that consider holistic faces, the proposed approach considers 3D facial components. It segments a 2D gallery face into components, reconstructs the 3D surface for each component, and recognizes a probe face by component features. The segmentation is based on the landmarks located by a hierarchical algorithm that combines the Faster R-CNN for face detection and the Reduced Tree Structured Model for landmark localization. The core part of the CNN-based approach is a revised VGG network. We study the performances with different settings on the training set, including the synthesized data from 3D reconstruction, the real-life data from an in-the-wild database, and both types of data combined. We investigate the performances of the network when it is employed as a classifier or designed as a feature extractor. The two recognition approaches and the fast landmark localization are evaluated in extensive experiments, and compared to state-of-the-art methods to demonstrate their efficacy.

%
\end{abstract}

\vspace{-1mm}
\section{Introduction}
\label{intro}
Cross-pose face recognition is generally handled by 2D or 3D based approaches. 
The 2D-based approaches require a training set from which the cross-pose characteristics can be learned and applied for recognition \cite{Heisele07, Lee13}. The 3D-based approaches are generally built on the holistic 3D surface reconstruction of a 2D face \cite{Prabhu11, Asthana11, HeoS12, 14_TIFS_Ali, _15_TIP_MASM}. 
Two approaches are proposed in this study, one is based 3D and the other is based on the Convolutional Neural Network (CNN). The 3D-based one differs from most 3D approaches in that it is built on the 3D reconstruction of facial components, instead of the whole face. Although component-based methods are popular in 2D approaches \cite{Heisele07, Lee13}, component-based 3D approaches for recognition are rarely seen.

In many 3D approaches, facial landmarks are exploited to align a 2D face to a 3D model. Various landmark detection approaches have been used, for example, the Active Appearance Model (AAM) \cite{Prabhu11, Asthana11}, Active Shape Model (ASM) \cite{14_TIFS_Ali, _15_TIP_MASM, yi2013towards}, Constrained Local Model (CLM) \cite{_15_TIFS_clm} and Tree Structured Model (TSM) \cite{14_TIFS_Ali}. Among these approaches, we consider the TSM the most advantageous because it offers 
two unique characteristics: 1) It detects faces and locates the landmarks simultaneously in one unified model; and 2) It can detect landmarks in a wider pose range, e.g., up to $90^o$ in yaw, whereas many other approaches only cover up to $45^o$, where both eyes are visible.
The disadvantage of the TSM is its sluggish runtime speed, which makes it impractical when handling real applications. We propose the Fast Hierarchical Model (FHM) as a key module in our approach for tackling cross-pose recognition. The FHM takes the advantages of the state-of-the-art Faster R-CNN (Region-based Convolutional Neural Network) \cite{ren2015faster} and of an architecture-revised TSM, called RTSM (Regressive Tree Structured Model), for rapid cross-pose landmark detection.

The proposed 3D-based approach operates in three phases. In Phase 1, the FHM locates the landmarks required for face alignment and facial component segmentation. The segmented facial components are then reconstructed in Phase 2. 
In Phase 3, the reconstructed 3D components are exploited for tackling cross-pose recognition using an approach based on the Sparse Representation-based Classification (SRC).

The proposed CNN-based approach strongly depends on the data employed to train the network. In this study, we train the network by three categories of data: 1) The data synthesized from the 3D reconstruction as in Phase 2 of the aforementioned 3D-based approach; 2) The in-the-wild real-life face dataset; and 3) Both combined.  
%
\begin{figure*}[t]
\centering
\includegraphics[width=15.2cm]{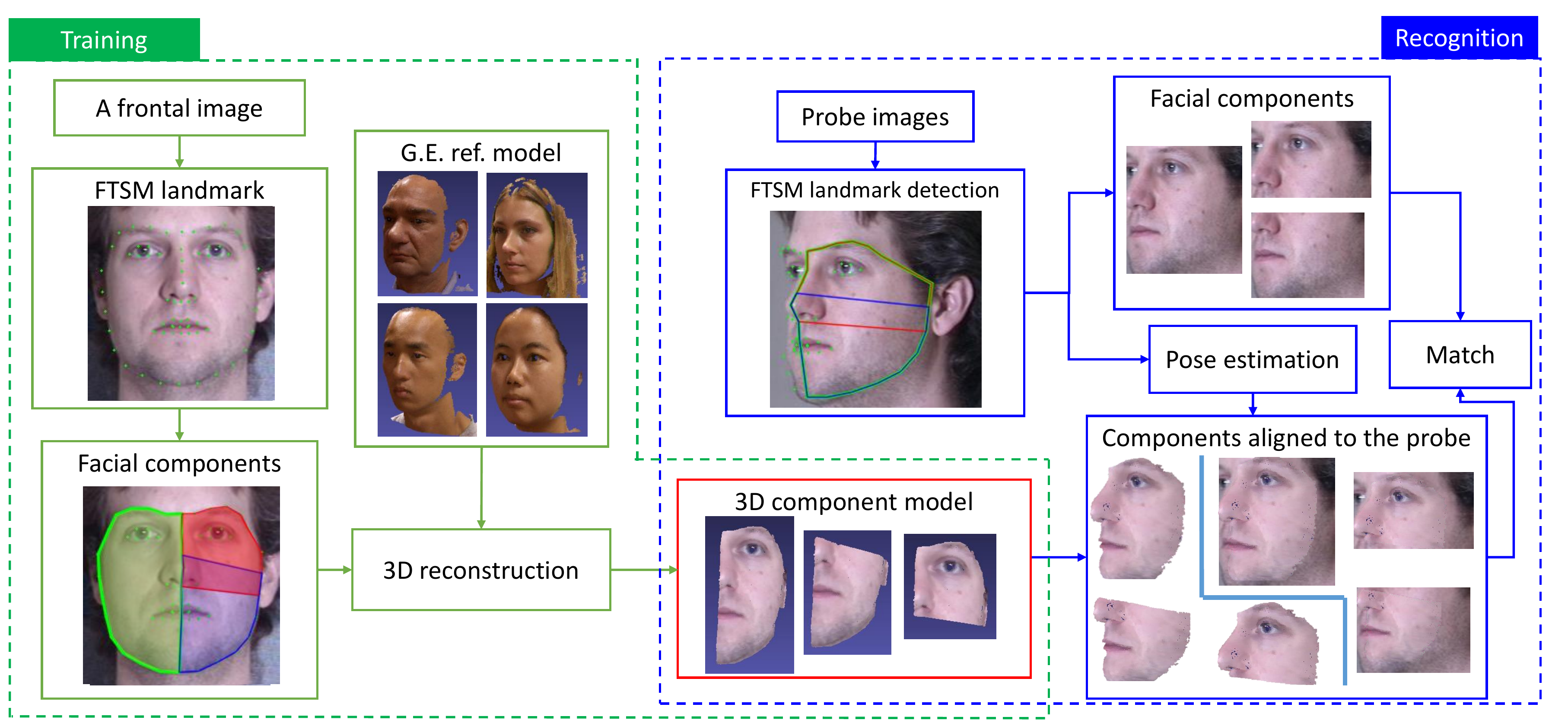}  
\caption{Workflow of the proposed method. At the training stage, each 2D face in the gallery is segmented into 6 components using the FHM landmarks (in Sec.~\ref{ptm}). These 6 components include 2 half faces and 4 quarter faces. For better visibility, the facial components boxes only show one half face on the left and two quarter faces on the right. Each component is reconstructed using an ethnic- and gender-oriented 3D reference model (in Sec.~\ref{recons}). At the recognition stage, the 2D image of each reconstructed component is aligned to the corresponding part of the query, and the match is determined by the approaches proposed in Sec.~\ref{recog}.}
\label{fig_flow}
\end{figure*}

The pros and cons of both approaches are highlighted and compared. The contributions of this work can be summarized as follows:
\begin{enumerate}
\item A 3D component based reconstruction is proposed and validated with state-of-the-art performance for cross-pose recognition. It is better in the computational cost and performance than most 3D approaches that consider holistic faces.
\item A CNN-based approach is proposed and validated with state-of-the-art performance as well. This study reveals the influences of training on real-life data, synthesized data and both types of data combined.
	\item The proposed FHM outperforms the TSM \cite{cvpr12_zhu} and many state-of-the-art approaches for landmark localization across large poses, and it retains the aforementioned advantages of TSM. 
	\item A comparison of the hand-crafted 3D component-based approach and a CNN-based solution is offered to show the individual advantages. The latter outperforms the former for cross-pose recognition only if the CNN is trained on a large collection of {\it in-the-wild} faces. If the large in-the-wild face collection is unavailable, the 3D component-based approach can be one of the most effective solutions.
\end{enumerate}


The rest of this paper is organized as follows. A review on previous works 
is given in Sec.~\ref{review}. The development of the FHM for landmark localization is presented in Sec.~\ref{ptm}, 
followed by the landmark-based 3D component reconstruction 
in Sec.~\ref{recons}, and the two proposed pipelines for recognition in Sec.~\ref{recog}. 
Experimental evaluations are presented in Sec.~\ref{sec_exp1} and \ref{sec_exp2}. Sec.~\ref{con} gives a conclusion to this study.

\section{Related Work}
\label{review}
%
%
%
\subsection{Review on Cross-Pose Recognition}
\label{sec_review_recog}
%
The method proposed by Asthana et al. \cite{Asthana11} exploits the View-based Active Appearance Model (VAAM) for landmark detection and support vector regression for pose normalization. 
However, because of the pose limitation of the VAAM, the method can handle poses up $45^\circ$ only.
The Generic Elastic Model (GEM) \cite{Prabhu11} presumes that the depth of a gallery face can be accurately reconstructed by a generic depth map, with 2D dense meshes built on the MASM (Modified Active Shape Model) landmarks for both the gallery face and the generic model. 
The GEM has been improved later in an extension work \cite{HeoS12}. Although the performance on the MPIE in the extension appears better than that of the original GEM \cite{Prabhu11}, both ignore the performance for large rotation, i.e., yaw angle $\geq60^\circ$ (the tests in \cite{HeoS12} only covers up to $30^\circ$).
The GEM has been used in several of the latest studies. The method proposed in \cite{14_TPAMI_Sparse} is built on the GEM, and it extracts sparse features by using subspace modeling and $l_1$-minimization to induce pose-tolerance. 
However, the experiment on MPIE is again limited to $\pm45^\circ$. A latest approach \cite{_15_TIP_MASM} uses the GEM and extracts Walsh local binary patterns from the periocular region as the features. The coupled max-pooling kernel class-dependence feature analysis (CoMax-KCFA) is harnessed to learn the pose-invariant subspaces for classification. This method outperforms that of \cite{Prabhu11} on MPIE, but again the pose is still limited to $60^\circ$.
%
%
Arguing that many 3D face models based on the Lambertian assumption ignore specular and diffuse reflection, a Heterogeneous Specular and Diffuse (HSD) approach with 3D surface approximation is proposed in~\cite{Zhang12}. This work considers the spatial variations of specular and diffuse reflections over a face, and it is experimentally shown effective for handling the extreme poses in the PIE database \cite{PIE}.
Nevertheless, the requirement that multiple frontal images with various illumination conditions are needed for the HSD surface approximation substantially impedes its practical application.

The review given above reveals the fact that most 3D-based methods 
are holistic and rarely exploit facial components, especially when considering 3D reconstruction and pose normalization. 
This review also justifies the requirement of facial landmarks in 3D approaches. However, only limited work presents the details on the computation of the landmarks. One of the motivations for this study is to merge landmark localization into the pipeline of cross-pose face recognition.

The deep learning approaches have been revolutionized many areas in computer vision, including face recognition.
The structure proposed by Ding and Tao \cite{Changxing2015} is composed of a set of CNNs and a stacked auto-encoder (SAE). The CNNs, trained on the CASIA-WebFace \cite{DongYi2014}, are used to extract facial features, and the SAE compresses the features. It achieves 98.43\% verification rate on the LFW database \cite{LFW}. Trained on a private (FaceBook) database with 4.4 million faces, the DeepFace single gives 95.92\% and the ensemble of 3 networks attains 97.35\% on LFW \cite{taigman2014deepface}. Singe and ensemble networks are employed elsewhere. Combining 25 CNNs and each trained on a different facial region, the DeepID2 reaches 99.15\%, however, the best single CNN yields 95.43\% \cite{nips2014_Sun}. To alleviate the issues with deeper and more complex CNNs, Wu et al. propose a lightened CNN framework that achieves 98.13\% \cite{WuHS15}. Recently, the features extracted from a CNN, also trained on the CASIA-WebFace, are exploited with the joint Bayesian metric learning, yielding 97.15\% on LFW \cite{JCChen2016}. All these methods are compared with the proposed approach in Sec.\ref{sec_exp2cnn}.


\subsection{Review on Facial Landmark Localization}		
Fully automatic facial landmark localization is split into two phases, the first is face detection and the second is the landmark localization on the detected faces. Most works on landmark localization and face alignment only focus on the second phase, assuming that the locations of faces can be obtained by a face detector \cite{CVPR13_Asthana, Xuehan13, Shaoqing14, IJCV11_Saragih}. The popular Viola-Jones detector is used in \cite{Xuehan13, Shaoqing14}, and the TSM is used in \cite{CVPR13_Asthana}. However, the Viola-Jones detector cannot handle faces with large rotation, 
and the TSM detector is too slow to be able to handle practical applications. Another big issue with these and many other landmark localization approaches is that they can only handle $45^\circ$ in yaw, while the cross-pose recognition considers profile-to-profile poses, i.e., up to $90^\circ$.

The Constrained Local Models (CLMs) \cite{CVPR13_Asthana, Xuehan13, Shaoqing14} and TSMs \cite{cvpr12_zhu, IJCV05_Felzenszwalb} are among the most successful approaches for landmark localization.
The Discriminative Response Map Fitting (DRMF), proposed in \cite{CVPR13_Asthana}, exploits an discriminative regression-based approach for improving fitting accuracy. 
A group sparse learning approach, called the Cascaded Deformable Shape Model (CDSM), is proposed by Yu et al. \cite{cdsm} to select the most salient facial landmarks. 
The Supervised Descent Method (SDM) is proposed by Xiong and De la Torre \cite{Xuehan13} for minimizing a Non-linear Least Squares (NLS) function formed by the initial and target landmark locations. The SDM learns a sequence of descent directions that minimizes the mean of the NLS functions sampled from the training set. 
Using a local binary feature set and a locality principle for learning those features, the Regressing Local Binary Feature (RLBF) achieves cutting-edge landmark precision and computational efficiency \cite{Shaoqing14}.
However, as mentioned above, the models considered in these approaches and many others are designed to capture faces with poses in which both eyes are visible, and therefore, they can be of limited use for tackling cross-pose recognition.


Unlike most CLMs that consider landmark fitting on a detected face, the TSM solves face detection and landmark localization 
in a unified framework \cite{cvpr12_zhu}. 
However, the major disadvantage of the TSM is the heavy computation required for runtime detection, which substantially impedes its capacity for handling practical applications. We propose the FHM (Fast Hierarchical Model) for solving this speed issue, while improving the performance of the original TSM. 
%
%

A few CNN-based approaches for landmark localization also deserve our attention. 
Zhang et al. \cite{zhang2016learning} propose the Tasks-Constrained Deep Convolutional Network (TCDCN), which not only learns the inter-task correlation but also employs dynamic task coefficients to facilitate the multi-task optimization. 
A three-level cascaded CNN, proposed by Sun et al. \cite{sun2013deep}, first extracts global features by the first level for initializing the landmark localization, and refines the initial predictions by the next two levels. However, they only locate 5 sparse landmarks in the facial region without considering any landmarks on the contour.
The Cascade Multi-Channel CNN (CMC-CNN) \cite{hou2015facial} locates the landmarks by performing bottom-up detection and top-down correction via a cascade of CNNs. Both local features and global constraints are exploited for locating the landmarks. Note that these CNN-based approaches only consider poses $\le45^\circ$, and cannot handle cross-pose landmark detection. The TCDCN and CMC-CNN are compared with the proposed approach in Sec.\ref{sec_exp1}.

\section{Fast Hierarchical Model for Cross-Pose Landmark Detection}
\label{ptm}

The proposed FHM (Fast Hierarchical Model) is composed of the Faster R-CNN \cite{ren2015faster} for high-performance face detection and the RTSM (Regressive Tree Structured Model), an improved TSM, for cross-pose landmark localization.

\subsection{Faster R-CNN for Face Detection}
\label{subsec_Faster}

The Faster R-CNN \cite{ren2015faster} is composed of a Region Proposal Network (RPN) and a Fast R-CNN detector \cite{fastrcnn}, and both share the same convolutional layers. The RPN aims to generate region proposals on a given image.
%
To generate the region proposals, a sliding window is used to scan over the feature map at the last shared convolutional layer. Multiple region proposals, called {\it anchors}, with different scales and aspect ratios are generated at each sliding-windowed location. The feature captured by each anchor is fed into two sibling fully connected layers, a regression layer and a classification layer. The regression layer gives the coordinates of the anchor and the classification layer verifies whether the anchor is an object. 
%
%
One the other hand, the Fast R-CNN \cite{fastrcnn} aims at improving the R-CNN \cite{rcnn} and SPPnet (Spatial Pyramid Pooling Network) \cite{sppnet} in both speed and accuracy. It takes the entire image and a set of object proposals as input. The network first processes the whole image with several convolutional and max pooling layers to produce a conv feature map. A region of interest (RoI) pooling layer is then used to extracts a fixed-length feature vector from the conv feature map for each object proposal. Each feature vector is fed into a sequence of fully connected layers that branch into two sibling output layers: one that produces softmax probability estimates over object classes plus a catch-all {\it background} class and another layer that outputs four real-valued bounding-box positions for each object class. The Fast R-CNN is trained end-to-end with a multitask loss.


%
\begin{figure}[t]
\centering
\includegraphics[width=8.8cm]{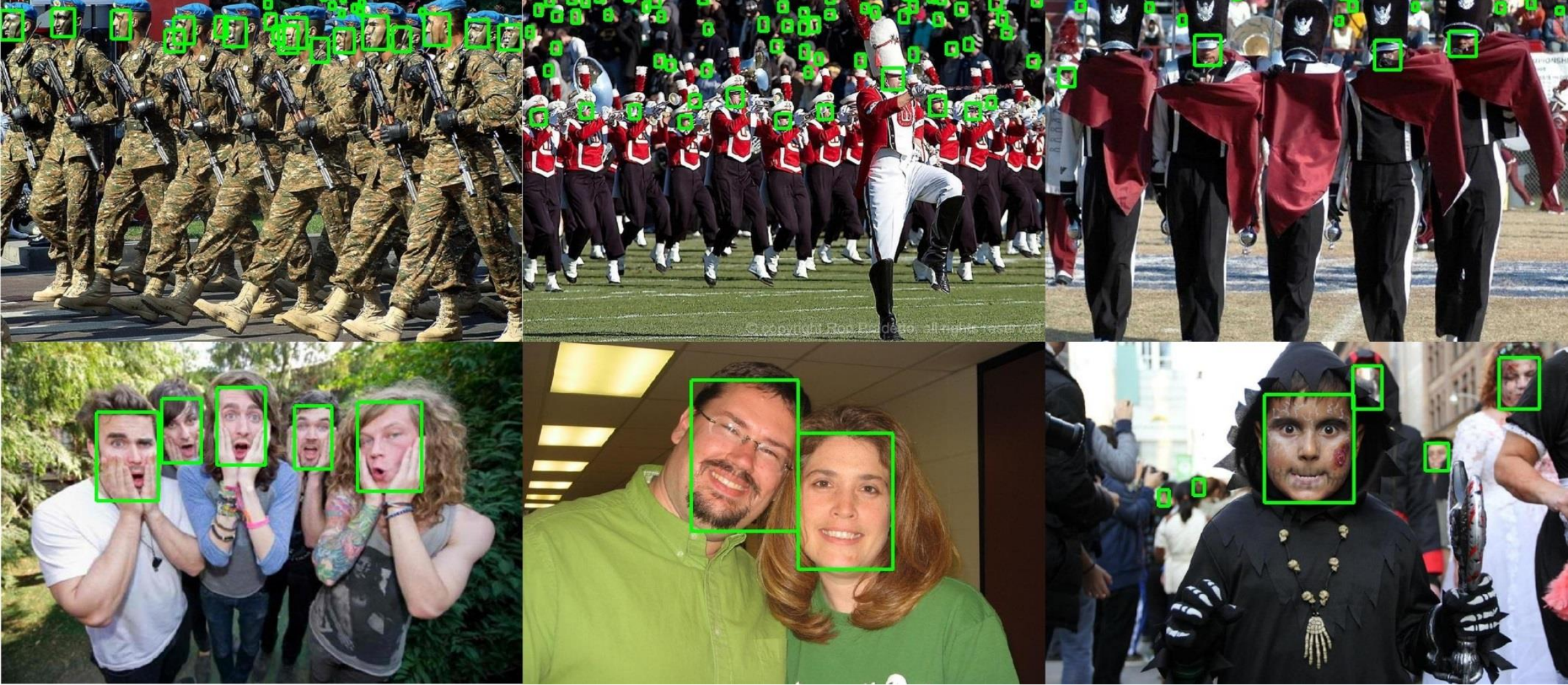}  
\vspace{-1mm}
\caption{Samples from the test set of the WIDER FACE \cite{yang2016wider} with detection bounding boxes given by the proposed Faster R-CNN face detector. Following the defined partition \cite{yang2016wider}, the WIDER FACE is split into training set and validation set with 16,106 images (199k faces) and test set with 16,097 images (194k faces).}
\label{widerface}
\end{figure}
The following 4-step training algorithm is exploited in \cite{ren2015faster} to learn shared features via alternating optimization.
\begin{enumerate}
	\item The RPN is initialized with an ImageNet-pre-trained model and trained end-to-end for performing the region proposal task. 
	To handle multi-scale localization, the pyramid of anchors is exploited for generating region proposals of different scales and aspect ratios.
	\item A Fast R-CNN for detection is also initialized by an ImageNet-pre-trained model, but trained using the region proposals generated in Step 1. At this point the RPN and the Fast R-CNN have not shared the convolutional layers yet.
	\item The RoIs given by the Fast R-CNN in Step 2 are used to reinitialize the RPN, which is trained on the same convolutional layers and only the layers unique to RPN are fine tuned. The two networks start to share convolutional layers from this phase on. 
	\item The Fast R-CNN is then reinitialize using the region proposals generated in Step 3, and trained on the shared convolutional layers with the unique layers of Fast R-CNN fine tuned. Step 3 and 4 can then be repeated on the same shared convolutional layers. The two networks finally form a unified network.
\end{enumerate}
We use the codes provided by the Ren et al. \cite{ren2015faster} on \url{https://github.com/ShaoqingRen/faster_rcnn}, and train the network using one of the latest large face databases, WIDER FACE \cite{yang2016wider}. The 393,703 labeled faces in the WIDER FACE reveal large variations in scale, pose, expression, skin color, illumination and occlusion conditions. Figure~\ref{widerface} shows a few samples from the test partition of the database with the detection bounding boxes given by the Faster R-CNN. 
Details of the experiments are reported in Sec.~\ref{sec_exp1}.


\subsection{Regressive Tree Structured Model (RTSM)}
\label{sec:review}
\begin{figure}[t]
\centering
\includegraphics[height=3.66cm]{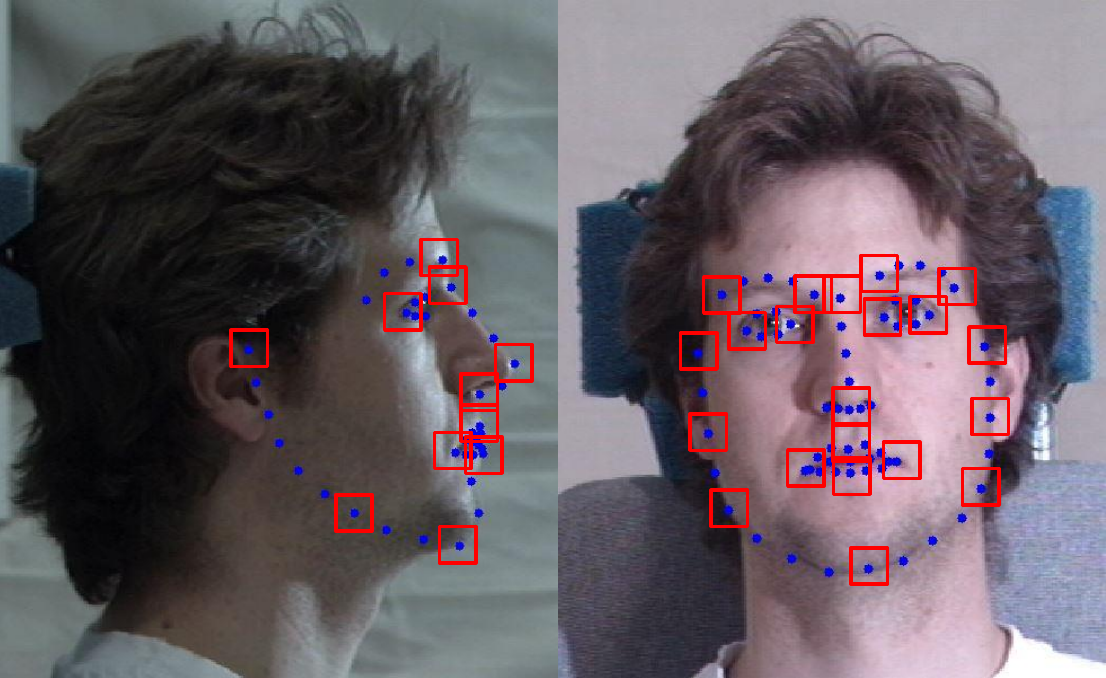}  
\caption{The blue dots enclosed by the red bounding boxes are the sparse landmarks to be located by the r-TSM, and the blue dots without the bounding boxes are the dense landmarks to be located by the BRM.}
\label{fig_landmarks}
\end{figure}

The RTSM is composed of two sub-models, the reduced Tree Structured Model (r-TSM) and the Bidirectional Regression Model (BRM). The r-TSM is developed based on the TSM but only with a sparse set of parts, and defined on samples of reduced dimension, making it more computationally efficient than the TSM. Given the sparse set of landmarks located by the r-TSM, the BRM generates the dense set of landmarks using the shape information learned from a training set. Figure~\ref{fig_landmarks} shows the sparse set of landmarks located by the r-TSM and the dense set located by the BRM.

As the r-TSM is built on the TSM, we give a brief introduction to the TSM first. A TSM $T$ consists of two components, $V$ and $E$, where $V$ is the set of parts, $E$ is the geometrical connection of the parts. 
Given an image $I$, the model can be expressed as a scoring function of a configuration ${\bf c}\in T = (V, E)$ in the following form \cite{cvpr12_zhu}:
\begin{eqnarray}
R(I,{\bf c}) = \sum_{k=0}^K {F_{k} \cdot \phi(I,s_k)} +
 \sum_{k,l\in E} b_{(k,l)} \cdot \rho(I,s_k,s_l) +\beta \label{model}
\end{eqnarray}
where $F_k$ is the filter for the $k$-th part associated with the local appearance feature $\phi(I,s_k)$ extracted from the part patch at location $s_k$ in $I$; $\{b_{(k,l)}\}$ is coined the spring parameter in \cite{cvpr12_zhu}, associated with the shape deformation feature $\rho(I,s_k,s_l)$, which is dependent on the locations of the parts at $s_k$ and $s_l$; and $\beta$ is a bias. When searching for the target object, one can maximize (\ref{model}) over all possible ${\bf c}$ so that the one with the most appropriate configuration ${\bf c}^*$ receives the highest score $R(I,{\bf c}^*)$.
%
The maximization of $R(I,{\bf c})$ can be performed via dynamic programming, which computes the message that Part $j$ passes to its parent Part $i$ as follows:
\begin{eqnarray}
n_{j} (s_i)&=&\max_{s_j}\left(g_{j} (s_j)+b_{(i,j)} \cdot \rho(I,s_i,s_j)\right)  \label{score2} \\
g_{j} (s_j)&=& F_{j} \cdot \phi(I,s_j)+\sum_{k\in K(j)}n_{k} (s_j)    \label{score1}
\end{eqnarray}
where $K(j)$ is the set of children of Part $j$. (\ref{score2}) computes the highest scoring location of its child Part $j$ for every location of Part $i$. (\ref{score1}) computes the local score of Part $j$, at all pixel locations $s_j$, by collecting messages from $K(j)$. When messages are passed to the root part ($j$ = 0), $g_{0} (s_0)$ gives the configuration with the best score for each root position. One can then use these root scores to generate multiple detections in $I$ by thresholding them and applying non-maximum suppression, and then backtrack to find the location and type of each part in each best-scored configuration by keeping track of the indices with each maximum.
%
%
In Zhu and Ramanan's study \cite{cvpr12_zhu} the histogram of oriented gradients (HoG) of the part patch is used as the part feature $\phi(I,s_j)$ at location $s_j$, and $\rho(I,s_k,s_l)=[dx_{k,l}\ dx_{k,l}^2 \ dy_{k,l}\ dy_{k,l}^2]$, where $dx_{k,l}$$=x_{s_k}$$-x_{s_l}$ and $dy_{k,l}=y_{s_k}-y_{s_l}$.
Given the dense set $L_d$ with 68 landmarks, the r-TSM is defined on $L_s$, a sparse subset selected from $L_d$, and the BRM captures the relationship between $L_s$ and $L_d$ by using shape traits. The r-TSM is trained on the downsized images which are the originals down-scaled with a scale factor $\sigma_h$. As the r-TSM has far fewer parts than the TSM and is defined on a reduced dimension, it is computationally more advantageous than the TSM.
Given the sparse landmark set $L_s$, the BRM exploits the Support Vector Regression (SVR) \cite{Smola03atutorial} to locate the rest of the dense landmarks using the shape characteristics learned from a training set.
The trained SVR, as written in the following form, can best describe the shape characteristics between the landmark locations $\mu_j\in L_s$ and the landmark locations $\nu_j\in [L_d-L_s]$. 
\begin{eqnarray}
\nu_j=\sum_{i=1}^M{(\alpha^*_i-\alpha_i)K(\mu_j,\mu_{i})+b_r} \label{svr}
\end{eqnarray}
where $\alpha^*_i$ and $\alpha_i$ are the Lagrange multipliers; $[\mu_i]$ are the support vectors, in terms of landmark locations: 
$b_r$ is the bias coefficient and $K(\cdot)$ is the kernel function. $K(\cdot)$ is chosen as the second-order polynomial as it leads to the best performance in our experiments.

To improve the accuracy of landmark locations, we select the landmarks with low location errors from the estimated $L_d^{(1)}$ as the reference input to the SVR and re-locate the landmarks to update the dense set to $L_d^{(2)}$. The selection of low-error landmarks is empirical and based on the landmark errors obtained in the training set. These low-error landmarks, denoted as $L_r$, are fixed at runtime. This processing is undertaken in a bidirectional way, as it starts using the r-TSM located $L_s$ to estimate $L_d^{(1)}$, and selects $L_r^{(1)}$ from $L_d^{(1)}$ to relocate the rest of landmarks and update the dense set to $L_d^{(2)}$, and then selects $L_r^{(2)}$ from $L_d^{(2)}$, and repeats until the difference between $L_d^{(i-1)}$ and $L_d^{(i)}$ falls below a predefined threshold.

When applying the TSM at runtime, one first computes the part feature pyramid on an image, and then searches for the faces by computing the distance transform between each search region and the TSM at all pyramid levels. This is a time-consuming processing as each search region must undergo the computations described previously, 
and it repeats for all pyramid levels to search for faces of all scales. It takes p-99, the fastest TSM, 2.6 secs to locate the landmarks on a face in MPIE; and 8.8 secs for p-1050, the most accurate TSM, to do the same. However, when using the proposed FHM on Cuda 6.5 (see Sec.~\ref{sec_exp1} for details), there can be up to 300 faces captured in an image in 65 milliseconds (ms). Each face is captured with a bounding box of an appropriate size, which makes the r-TSM focus on limited scales of faces only, substantially reducing the search time.

For comparison purpose, we design a simplified version of the r-TSM, called the coarse TSM (c-TSM), for face detection. Like the r-TSM, the c-TSM is also built on the sparse landmark set, 
as those shown in Figure~\ref{fig_landmarks}. However, the c-TSM is defined on the facial images half in scale of that defines the r-TSM. The training samples considered for the TSM \cite{cvpr12_zhu} are around $200^2$ pixels. We define the RTSM with scale factor $\sigma_h=0.5$, i.e., the r-TSM and BRM are trained on faces of $100^2$ pixels, and the c-TSM is trained on faces of $50^2$ pixels. 
Experiments on benchmark databases, as reported in Sec.~\ref{sec_exp1}, demonstrate the advantage of the proposed FHM over many contemporary approaches. The FHM landmarks are exploited in the following component-based reconstruction and recognition phases. 

\section{Landmark-based Facial Component Reconstruction}
\label{recons}
Each 2D face in the gallery can be decomposed into six component regions by using the FHM landmarks, as shown 
in Figure~\ref{fig_flow}. The six components are defined to meet two requirements: 1) Some components must be kept visible at large rotation; 2) Each component must enclose a sufficient amount of 3D surface curvature. The six component regions include two half faces and four quarter faces. 
For better visibility, the components in Figure~\ref{fig_flow} only show one half face on the left, and two quarter faces on the right. We have modified the approach proposed by Kemelmacher-Shlizerman and Basri \cite{KShlizerman11}, called KB reconstruction, for the 3D reconstruction of each component region. The reconstruction involves two steps: 1) Surface smoothing and surface parameter estimation for the reference model, and 2) The gender- and ethnicity-oriented 3D reconstruction of the facial components. 

\subsection{Surface Smoothing and Parameter Estimation for Reference Model}
\label{Ref_coeff}
As in many reconstruction methods that require 3D face models as bases or references \cite{Prabhu11, Asthana11, KShlizerman11}, our reconstruction also needs a 3D reference model for the initial values of the surface parameters, 
including the noise-free depth and surface normal.
We exploit the 3D scans from the FRGC database \cite{FRGC}, because the 3D scans offer plenty of 2D images good for estimating the albedo needed in the reconstruction. However the FRGC 3D face scans do not offer the aforementioned surface parameters, and we must estimate these surface parameters by using the raw depth map and texture image that come with the face scan. Additionally, because the raw depth map is often corrupted to some extent by measurement noise, we need to smooth the raw depth map before estimating the surface parameters. This step is not described explicitly in the aforementioned papers or in others, but it is considered an essential part of the reconstruction when using a raw 3D face scan as the reference model.

We revise the Moving Least Squares (MLS) \cite{AlexaS03} for smoothing $z_{r,0}$ the raw depth data of the reference model, so that the measurement noise in $z_{r,0}$ can be removed and the smoothed surface $z_{r}$ can best approximate $z_{r,0}$. We use the FHM landmarks to define triangle mesh to decompose the face into hundreds of triangular regions in the form of point cloud, and consider each triangular region a point cloud subset. 
Given a subset ${\bf P}_k$$=$ $\{{\vec{p}}_i\}_{i=1,\cdots,N_k}\in z_{r,0}$, the goal is to determine a novel set of points, ${\bf R}_k$$=$ $\{{\vec{r}}_i\}_{i=1,\cdots,N_k}$, on a low-order polynomial that minimizes the distance between ${\bf P}_k$ and ${\bf R}_k$. The smoothed and noise-free surface $z_r$ can then be obtained from $\{{\bf R}_k\}_{\forall k}$. Our revised MLS includes the following steps: 
\begin{enumerate}
\item Use ${\bf P}_k$ to determine a local plane $H_0$ with origin ${\vec{q}}_0$ and normal ${\vec{n}}_0$ so that the following weighted sum can be computed, 
\begin{equation}
\sum _{i=1}^{N_k} \left( u_0(x_i,y_i)-\mu_{i,0} \right)^2  \phi \left( \left\| \vec{p}_i-\vec{q} _0\right\|  \right)
\label{MLS}
\end{equation}
where $u_0(x_i,y_i)$ is the distance of $\vec{r}_i$ to $H_0$ with the location of its projection onto $H_0$ given by $(x_i,y_i)$; $\mu_{i,0}$ is the distance of $\vec{p}_i$ to $H_0$, i.e., $\mu_{i,0}=\vec{n}_0\cdot(\vec{p}_i-\vec{q}_0)$; and $\phi(\cdot)$ is a Gaussian function by which the points closer to $\vec{q}_0$ are weighted more. Assuming that ${\bf R}_k$ are described by a low-order polynomial in terms of the coordinates $(x_i,y_i)$ on $H_0$, i.e., $\vec{r}_i = f(x_i,y_i|\Lambda_0)$ and $u(x_i,y_i)={\vec{n}_0}\cdot(f(x_i,y_i|\Lambda_0)-\vec{q}_0)$, where $f(x_i,y_i|\Lambda_0)$ is the low-order polynomial surface with parameter $\Lambda_0$ that defines the local geometry of ${\bf R}_k$.
\item As $H_0$ can be uniquely defined given ${\vec{q}}_0$ and ${\vec{n}}_0$, we can change them to ${\vec{q}}_1$ and ${\vec{n}}_1$ and thereby obtain a novel plane $H_1$. Given that the order of the polynomial $f(x_i,y_i|\Lambda_0)$ is fixed (so that the number of parameters of $f(x_i,y_i|\Lambda_0)$ is fixed), a parameter estimation problem can be defined as the minimization of the weighted sum:
%
\begin{equation}
\hspace{-2mm}\Lambda^*_k,\vec{n}^*_k,\vec{q}^*_k = \argmin_{\Lambda,\vec{n},\vec{q}}\sum _{i=1}^{N_k} \left( u(x_i,y_i)-\mu_i \right)^2  \phi \left( \left\| \vec{p}_i-\vec{q} \right\|  \right)
\label{MLS_loop}
\end{equation}
\end{enumerate}
The above processing can be repeated on other subsets $\{{\bf P}_k\}_{\forall k}$ for estimating $\{\Lambda_k,\vec{n}_k,\vec{q}_k\}_{\forall k}$ and $\{{\bf R}_k\}_{\forall k}$.
Based on our experiments, we find that the minimum principal component extracted from ${\bf P}_k$ offers a good initial estimate for $\vec{n}_0$, and the centroid of ${\bf P}_k$ can be considered as $\vec{q}_0$.
%
%
Following the above approach, the surface normal $\vec{n}_r$ can be obtained from the estimated polynomials $f(x_i,y_i|\Lambda_k)$.


\subsection{3D Reconstruction of Facial Components using Ethnicity- and Gender-Oriented Reference Model}
\label{sec_3d_recons}
%
The differences between the proposed algorithm and the original KB reconstruction \cite{KShlizerman11} are summarized as follows:
\begin{itemize}
	\item The proposed approach exploits the 2D frontal images available in the FRGC database for approximating the albedo of the reference model, unlike the KB reconstruction which considers the albedo as an additional constraint;
		\item The proposed approach exploits the ethnicity- and gender-oriented reference model, which gives better performance in facial depth reconstruction than the generic reference model considered in the KB reconstruction \cite{KShlizerman11}. 
		\item The minimization conducted in the KB reconstruction concerns the whole face. Hence, it works well on low frequency (or smooth) regions such as cheeks, but produces relatively large errors on the high frequency regions such as eyes, nose and mouth. The errors on the high frequency regions are evened out with errors on the low frequency regions when minimizing the overall depth error, thus making further error reduction difficult. The proposed scheme focuses on the errors at the component levels. As components reveal better discriminating features, the scheme leads to better recognition.  
	\item As only components are considered, the proposed scheme comes with a lower computational and storage cost. 
\end{itemize}

Our approach starts from the FHM-based face alignment. Each component segmented from a 2D gallery face is aligned to the same component as segmented from the 3D reference model by using the FHM landmarks. Given the 2D component as the target $t(x,y)$ and the 3D component as the reference, and assuming that the face is Lambertian, the following formulation can be used to recursively estimate the surface reflectance $R(x,y)$, the depth $z(x,y)$ 
and the surface normal $\vec{n}(x,y)$ of the target.
\begin{equation}
\min_{\vec{l},\vec{z},\rho}
\int (t(x,y)-\rho(x,y)R(x,y))^2+\lambda_z(L_g \ast d_z)^2 dxdy
\label{opti_fun}
\end{equation}
where $\rho (x,y)$ is the surface albedo at pixel coordinates $(x,y)$; the reflectance $R(x,y)$ is the inner product of the lighting intensity $\vec{h}(x,y)$ and the surface normal $\vec{n}(x,y)$, i.e., $R(x,y)=\vec{h}(x,y) \cdot \vec{n}(x,y)$; $d_z$ is the difference between the unknown depth $z(x,y)$ and the reference depth $z_{r}(x,y)$, i.e., $d_z=z(x,y)-z_{r}(x,y)$. $L_g \ast$ denotes the convolution with the Laplacian of Gaussian (LoG); $\lambda_z$ is a 
constant. Applying LoG serves to locate large depth differences, 
and to force the minimization focused on the large-difference spots. The first term in (\ref{opti_fun}) describes the objective to be minimized, which is the difference between the target and the projection of the lighting $\vec{h}(x,y)$ that is cast on the surface at $(x,y)$. This term cannot be solved alone. Therefore, the second term is added as a constraint imposed on the depth. The problem is then formulated as a constrained minimization, and the solution can be obtained in an iterative way.

With a few assumptions \cite{KShlizerman11}, the reflectance $R(x,y)$ can be approximated using the lighting coefficient vector $\vec{l}(x,y)$ and spherical harmonics $\vec{Y}(\vec{n})$, i.e., $R(x,y)$$\approx \vec{l}$$ \cdot $$\vec{Y}(\vec{n})$.
For simplicity of notation, the coordinates $(x,y)$ are dropped in the rest of the paper. Therefore, for example, $\vec{n}(x,y)$ is written as $\vec{n}$. The difference between $\vec{h} \cdot \vec{n}$ and $\vec{l} \cdot\vec{Y}(\vec{n})$ is that the lighting intensity and direction are both merged into $\vec{h}$ in the former, but in the latter they are expanded in terms of the spherical harmonics $\vec{Y}(\vec{n})$.

\begin{figure}[t]
\centering
\includegraphics[width=8cm]{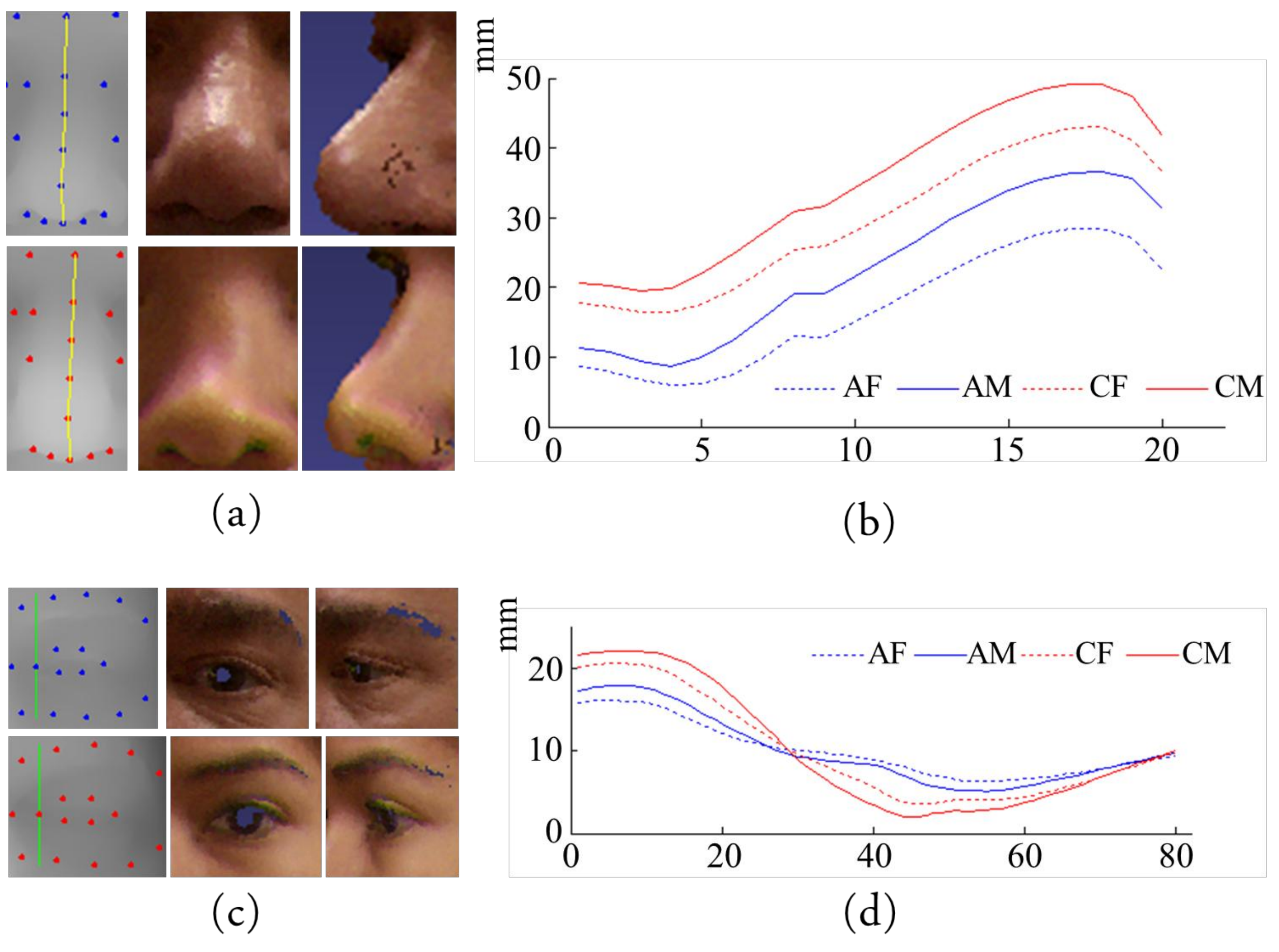}  
\caption{(a) shows the depth map and depth variation at the nose, viewed in the front and 60$^\circ$ from the side. (b) shows the mean depth (in mm) along the landmarks on the nose (in yellow line). (c) shows the depth map and depth variation along the inner side of the eye socket, and (d) shows the mean depth (in mm) along the inner side of the eye socket (the green lines in (c)). Results obtained based on 16 samples arbitrarily selected from each of ethnicity-gender groups in the FRGC 3D scan subset. The depth of the deepest eye corner is taken as the ground level.}
\label{fig_depth_samples}
\end{figure}

To study the effects made by reference models of different races and genders, four ethnicity-gender (E+G) groups with 16 samples in each group are arbitrarily selected from the 3D subset of the FRGC database. The four groups are Caucasian male (CM) and female (CF), and Asian male (AM) and female (AF). The faces in each group are aligned using the FHM landmarks, and averaged in all dimensions. The average is considered as the reference model contributed by each group. 
Figure \ref{fig_depth_samples} shows the depths along the landmarks on the nose and along the inner side of the eye socket. 
The Caucasians reveal higher nose and larger depth variation than the Asians. This depth difference is primarily caused by ethnicity rather than gender.

The reconstruction processes the minimization in (\ref{opti_fun}) by first solving for the spherical harmonic coefficients $\vec{l}(x,y)$ by using the reference $\vec{n}_r$ and assuming $z=z_r$. With $\vec{l}$ solved, we can recompute $z(x,y)$, then update $\rho(x,y)$, and next update  
{$\vec{n}$} as ${\vec{n}}=(p,q,-1)^T/\sqrt{(p^2+q^2+1)}$, 
where $p=\partial{z}/\partial{x}$ and $p=\partial{z}/\partial{y}$. The spherical harmonic coefficients $\vec{l}$ can then be updated. This process is repeated until the estimates converge.
%

\section{Recognition using SRC and CNN}
\label{recog}
Two approaches are proposed, one exploits the Sparse Representation-based Classification (SRC) and the other explores the Convolutional Neural Network (CNN). The former is often associated with hand-crafted features and the latter extract {\it deep features} from a large dataset. 

\subsection{Component-based Recognition with SRC}
\label{recog1}

When a query image is given, its pose is first estimated using the FHM landmarks,
and its facial components are cropped accordingly. The reconstructed component models of the gallery set are then rotated to the estimated pose of the query so that the 2D projection of the reconstructed components can be aligned with the query components. The illumination normalization proposed by Tan and Triggs \cite{10_TIP_TT} is applied on both the query and the aligned model for removing unbalanced illumination. 

The SRC is shown to be effective for handling illumination, expression and occlusion \cite{WrightYGSM09, ESRC}, but is rarely attempted for tackling pose and facial components. Therefore it is explored in this study. 
Given a set of the 2D projections of the aligned reconstructed components of the faces in the gallery, denoted as $M=[m_1, m_2, \cdots, m_k]$, and the same components of a query $q$, the core part of the SRC solves for the linear representation of $q$ in the span of $A$, where $A=[a_1, a_2, \cdots, a_k]$ is a matrix with its column $a_i$ being a feature vector extracted from $m_i$. We can therefore write $q^*=Ar^*+\mu^*$, where $r$ is a sparse vector and $\mu$ is a noise with bounded energy, i.e., $||\mu||_2<\epsilon$. Following the rules in compressing sensing \cite{WrightYGSM09}, $r^*$ can be obtained by solving the following $l_1$-minimization:
\begin{equation}
\hat{r^*}=\argmin ||r||_1, \ \ {\rm subject \ to} \ ||q-Ar||_2\le\epsilon  \label{eqn_L1_min}
\end{equation}
%

A comprehensive discussion on the solutions for the aforementioned $l_1$-minimization is given in \cite{YangSGM10}, where five fast algorithms are evaluated. 
We select the best two algorithms, the TNIP (Truncated Newton Interior-Point) and Homotopy, to evaluate their performances against pose variations. 
%

The TNIP exploits gradient projection (GP) and searches for the sparse vector $r$ along certain gradient direction with fast convergence. It reformulates (\ref{eqn_L1_min}) into the following form:
\begin{equation}
\hat{r^*}=\argmin_r \frac{1}{2}||q-Ar||_2^2 + \lambda||r||_1   \label{eqn_L1_TNIP}
\end{equation}
where $\lambda$ is the Lagrange multiplier. Such a formulation enables the solution using quadratic programming.

The Homotopy finds a solution path $X_h$ that varies with $\lambda$, 
\begin{equation}
X_h = \{ r^*_\lambda : \lambda \in [0, \infty)  \}   \label{eqn_homo}
\end{equation}
%
%
When $\lambda\rightarrow \infty $, $r^*_\lambda=0$, and when $\lambda\rightarrow 0$, $r^*_\lambda$ converges to the solution.  The Homotopy algorithm considers that the objective function in (\ref{eqn_L1_TNIP}) changes as a homotopy from the $l_2$ constraint to the $l_1$ objective as $\lambda$ decreases.
%
We use the SparseLab Toolbox \url{http://sparselab.stanford.edu} for solving (\ref{eqn_L1_min}).
%
\begin{table}[t]
\begin{center}
\caption{Architecture of the VGG network exploited. Each of first 5 blocks consists of 2 or 3 convolution layers and one max pooling layer. Block 6 is for dropout processing. Blocks 7 and 8 are with single fully-connected layers Fc7 and Fc8, with output dimension 512 and $N_c$, respectively. $N_c$ is the number of classes/subjects to identify}
\label{cnn}
\vspace{2mm}
\footnotesize
\begin{tabular}{|c|c|c|c|c|}
  \hline
	\multirow{2}{*}{Name} & \multirow{2}{*}{Type} & Output & Filter Size/ & \multirow{2}{*}{ReLu} \\
	 &  & Number & stride/pad & \\
		\hline
	Conv1\_1 & Conv & 64 & 3/1/1 & Yes \\
		\hline
	Conv1\_2 & Conv & 128 & 3/1/1 & Yes \\
		\hline
	Pool1 & Max pool & N/A & 2/2/0 & No \\
		\hline
	Conv2\_1 & Conv & 64 & 3/1/1 & Yes \\
		\hline
	Conv2\_2 & Conv & 128 & 3/1/1 & Yes \\
		\hline
	Pool2 & Max pool & N/A & 2/2/0 & No \\
		\hline
	Conv3\_1 & Conv & 128 & 3/1/1 & Yes \\
		\hline
	Conv3\_2 & Conv & 128 & 3/1/1 & Yes \\
		\hline	
	Conv3\_3 & Conv & 128 & 3/1/1 & Yes \\
		\hline
	Pool3 & Max pool & N/A & 2/2/0 & No \\ 	
		\hline
	Conv4\_1 & Conv & 256 & 3/1/1 & Yes \\
		\hline
	Conv4\_2 & Conv & 256 & 3/1/1 & Yes \\
		\hline	
	Conv4\_3 & Conv & 256 & 3/1/1 & Yes \\
		\hline
	Pool4 & Max pool & N/A & 2/2/0 & No \\ 	
		\hline
	Conv5\_1 & Conv & 256 & 3/1/1 & Yes \\
		\hline
	Conv5\_2 & Conv & 256 & 3/1/1 & Yes \\
		\hline	
	Conv5\_3 & Conv & 256 & 3/1/1 & Yes \\
		\hline
	Pool5 & Max pool & N/A & 2/2/0 & No \\ 	
		\hline	
	Drop6 & Dropout & N/A & N/A & No \\ 	
		\hline
	Fc7 & Fully conn & 512 & N/A & Yes \\ 	
		\hline
	Fc8 & Fully conn & $N_c$ & N/A & N/A \\ 	
		\hline
	Output & Softmax & $N_c$ & N/A & N/A \\ 	
		\hline
\end{tabular}
\end{center}
\end{table}

We compare different features, including pixel intensities, LBP (Local Binary Patterns) and Gabor features (obtained by the Gabor transform).
The recognition by each component is determined by the Rank-1 result, and the overall recognition is determined by the votes from all components. When the votes are tied, as is observed in some cases with four components, the Rank-2 results are taken into account. Our comparison study shows that the Homotopy with Gabor features delivers the best overall performance. We therefore only report the results with such settings in Sec.\ref{sec_exp2}.

\subsection{CNN-based Recognition}
\label{recog2}
The CNN exploited in this study is a revised version of the VGG network as it shows state-of-the-art performance for object recognition \cite{simonyan2014very}. We take the network from the open-source deep learning toolkit Caffe \cite{jia2014caffe}, and modify the input layer and the fully connected layers so that they fit the needed input and output settings. The network has 9 blocks and each block is composed of 1$\sim$3 layers, and there are 22 layers overall, as summarized in Table~\ref{cnn}. The input to the network is a $128\times 128$ gray-scaled image. Each input image is filtered by a stack of convolutional layers, operated with a 3×3 receptive field. The convolution stride is fixed to 1 pixel. The spatial padding of the convolutional layer input aims to keep the spatial resolution the same after the convolution, and is thus chosen as 1 pixel for the 3×3 convolutional layers. Five max-pooling layers follow the last convolutional layers in the first five blocks for pooling the outputs, which are the inputs to the next blocks. Max-pooling is carried out by a 2×2 window, with stride 2. All hidden layers are equipped with the rectified linear units (ReLUs) as the activation functions. To reduce overfitting, the sixth block, Drop6 in Table~\ref{cnn}, is used for dropout operation \cite{Hinton2012}, with dropout rate 0.5. The next two blocks, Fc7 and Fc8, are fully connected layers. They are the same as the convolutional layer, except that the size of the filters matches the size of the input data. The output dimension of Fc7 is 512, and that of Fc is the same as the number of classes to identify, which is denoted as $N_c$. The last output block consists of $N_c$ channels and each channel is with a softmax activation function.


The following two approaches are implemented to utilize the revised VGG network.
\begin{enumerate}
	\item The network is considered as a classifier trained to recognize multiple classes of faces, where each class corresponds to a subject. The identity of a query face presented to the network is determined by the softmax activation at the output. This approach is called Activation-at-Output (AO), which is appropriate for solving multi-classification, such as face identification.
	\item The network is considered as a feature extractor for extracting facial features. The extracted features can be exploited with classifiers 
	for handling verification or identification. We take the output of the first fully connected layer Fc7 as the feature, and call this approach Fully-Connected Feature (FCF).
\end{enumerate}

We train the VGG network on the recently released CASIA-WebFace database \cite{DongYi2014}, which has 494,414 facial images of 10,575 subjects collected in the wild. 
As the images in the CASIA-WebFace are collected in a semi-automatic way, part of the images are of poor quality or mislabeled. We remove most of these poor-quality and mislabeled images manually, and run the FHM to locate the landmarks on the remaining 381,975 images of 8,984 subjects (it would have taken much longer time if we used the TSM). $N_c$ in our experiments is thus chosen as 8,984.

Given the faces with landmarks, we compare the CNN performances for face identification across three settings: 1) with original face images, 2) with faces scaled to the landmarks, and 3) with faces scaled and aligned to the landmarks. 
Using a 5-fold cross-validation test, we have verified that the third settings yield the best identification rate at $83.51\%$, which is better than $76.53\%$ obtained by using the multimodal CNN proposed by Ding and Tao \cite{Changxing2015}.

The above study has verified the capacity of the VGG network trained on a large in-the-wild face database. This verification motivates us to investigate the following settings for handling the cross-pose recognition.
\begin{enumerate}
	\item Use the 3D reconstructed models of the faces in the gallery to generate a large set of synthetic 2D faces of all poses, and use this synthetic dataset to train the network, and then test the network by using the real 2D faces of non-frontal poses.
	\item Use the above trained-on-CASIA-WebFace network to test its performance on non-frontal faces.
	\item Use the above synthetic 2D faces combined with the CASIA-WebFace to train the network, and test on non-frontal faces.
\end{enumerate}

Note that the datasets for training the VGG network are too large and too complex to be segmented into facial components. Therefore, we only consider holistic faces when exploring the CNN-based pipeline. The FHM landmark localization is again considered a vital component for such a solution as it is required for the scale normalization and alignment of the training and testing samples in high speed.

\section{Experiments on FHM Landmark Localization}
\label{sec_exp1}
In this section, we present the experiments on the FHM landmark localization, and the experiments on cross-pose recognition are given in the next section.
All experiments were run on Cuda 6.5 with Caffe Matlab wrapper upon a Windows-7 PC with i7 (3.4GHz) CPU, RAM 16GB and Titan X GPU. 

\subsection{Experimental Setup}
\label{sec_exp1s}
%
The FHM has two modules, the Faster R-CNN for face detection and the RTSM for the landmark localization. We keep the same settings for the Faster R-CNN as those adopted in \cite{ren2015faster}, in which the network is employed for object detection. The codes for the Faster R-CNN were downloaded from \url{https://github.com/ShaoqingRen/faster_rcnn}. 
The Faster R-CNN is trained on the WIDER FACE database \cite{yang2016wider}, which is composed of 32,203 images and 393,703 labeled faces. These faces demonstrate a large degree of variability in scale, pose, expression and occlusion. This database also contains many extreme cases, such as faces smaller than $12\times12$ or poses with views from above or behind the faces. Human visual detection of these extreme cases often requires other body parts as priors, and they can be hardly detectable to human eyes when the body parts are removed or occluded. According to our experiments, the inclusion of these extreme samples in the training set degrades the detection performance, we run a series of cross-validation tests to measure the performance with and without these extreme samples in the training set. We find that the training set without faces $<12\times12$ yields a lower false positive and better Average Precision (AP).

Two networks, the ZF \cite{zeiler2014visualizing} and VGG-16 \cite{simonyan2014very}, are considered in our experiments for sharing the convolution layers in the Faster R-CNN pipeline. The ZF network has 5 sharable convolutional layers, and the VGG-16 network has 13 sharable convolutional layers. The training of ZF took almost 12 hours, but it took 32 hours to train the VGG-16. However, the latter gives AP $97.92$ and the former gives $97.06$, with runtime speed 65 ms (ZF) v.s. 169 ms (VGG) using 300 proposals per image.

The face detector reported in the work \cite{JiangL16a} also uses the Faster R-CNN trained on the WIDER FACE. The differences are threefold: 1) We remove tiny ($<12\times12$) faces, as mentioned above, as the features learned from tiny faces degrade the performance; 2) We compared the performances between VGG and ZF networks used as the sharing convolution layers, but they only use VGG; 3) They modified some original settings in \cite{ren2015faster} but we follow most of the original settings.
\begin{figure}[t]
\centering
\includegraphics[width=7.8cm]{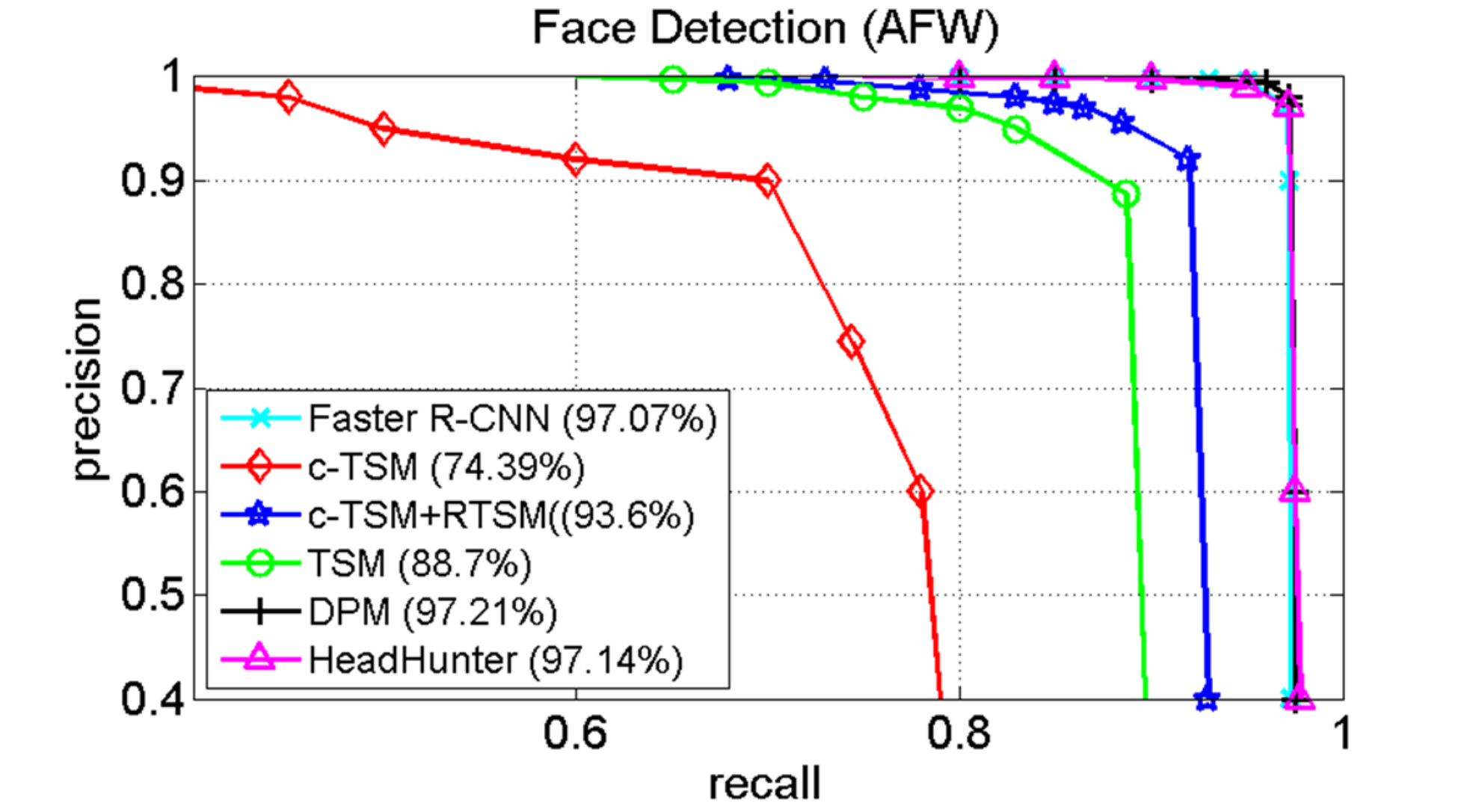}  
\vspace{-1mm}
\caption{Precision-recall rates on the AFW dataset with Average Precision (AP) in the parentheses. TSM is the p-1050 model \cite{cvpr12_zhu}. The DPM and HeadHunter are both from Mathias et al. \cite{Mathias2014Eccv}}
\label{ap_afw}
\end{figure}
\begin{figure}[t]
\centering
\includegraphics[width=7.8cm]{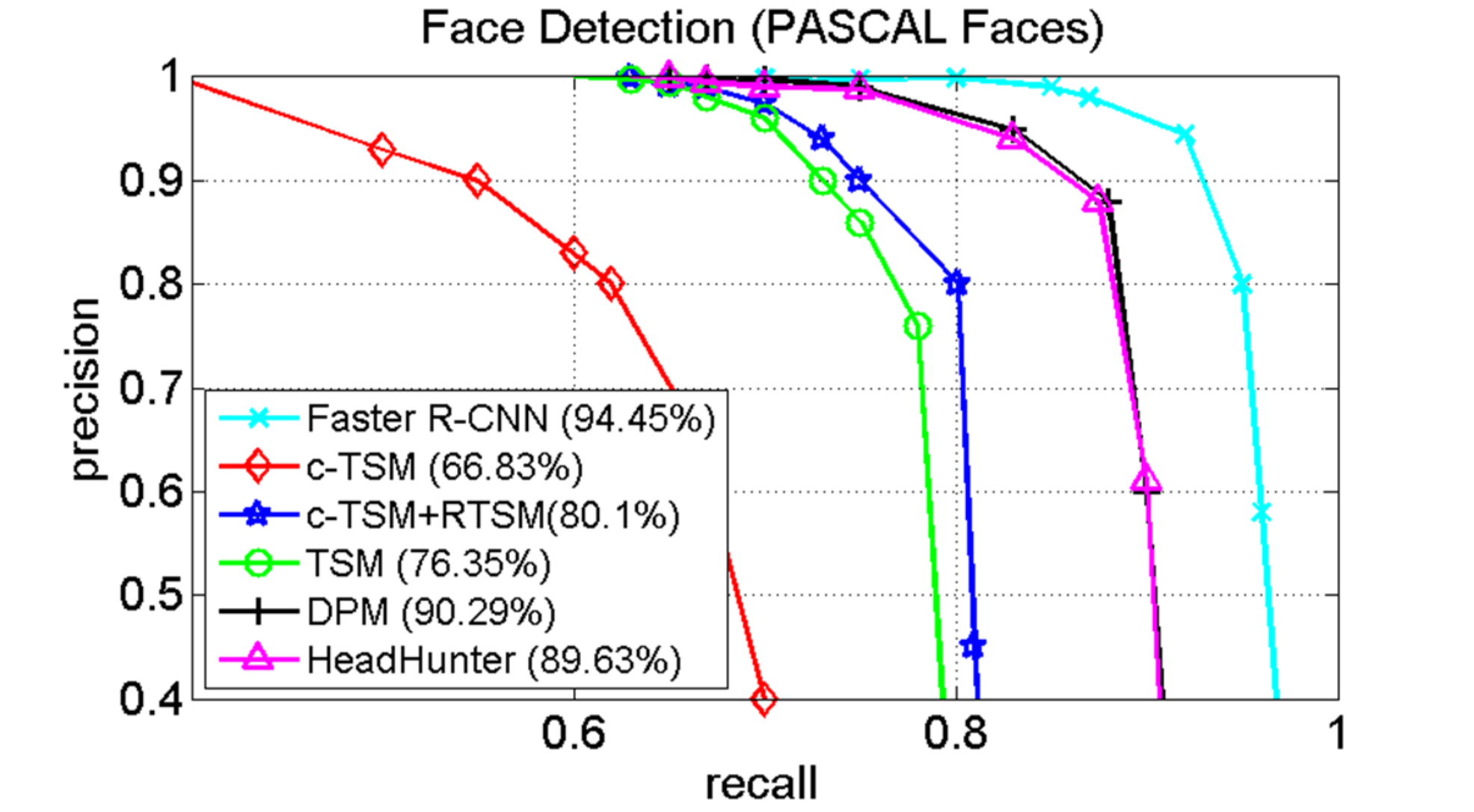}  
\vspace{-1mm}
\caption{Precision-recall rates on the PASCAL FACE dataset with Average Precision (AP) in the parentheses.}
\label{ap_pascal}
\end{figure}
\begin{table*}[t]
\begin{center}
\footnotesize{
\caption{{Landmark accuracy and localization-only runtime speeds.} {The location error is normalized to either inter-pupil distance ($\le45^o$) or eye-to-mouth ($>45^o$) distance, in terms of $\%$.} {($\cdot$) indicates the time including face detection and landmark localization. $^*$ indicates running on Matlab}}
\label{tab_LM}
\begin{tabular}{|c|c|c|c|c|c|c|c|c|}
\hline & \multicolumn{2}{|c|}{AFW} & \multicolumn{3}{c|}{300-W} & \multirow{2}{1cm}{Overall} & \multirow{2}{2cm}{Runtime/Face \\ in MPIE (ms)} & \multirow{2}{2cm}{Runtime/Face \\ in AFW (ms)} \\
\cline{1-6}
Method &   $ > 45^\circ$ & $\le 45^\circ$ & Common & Challenge & Full Set &  & &\\
\hline
SDM \cite{Xuehan13} & 39.20 & 8.80 & 5.57 & 15.40 & 7.50 & 18.50 & 25 & 29 \\
\hline
RLBF \cite{Shaoqing14} & 38.70 & 8.61 & 4.95 & 11.98 & 6.32 & 17.88 & 10 & 11 \\
\hline
DRMF \cite{CVPR13_Asthana} & - & 12.90 & 6.65 & 19.79 & 9.22 & - & 0.9k* & 1.1k*\\
\hline
TSM \cite{cvpr12_zhu} & 7.11 & 14.30 & 8.22 & 18.33 & 10.20 & 10.54 & (8.8k), (25.6k)* & (24.9k), (71.2k)* \\
\hline
CDSM \cite{cdsm} & 6.62 & 11.79 & - & - & - & - & - & (5.8k)*  \\
\hline
CMC-CNN \cite{hou2015facial} & - & - & 4.91 & 12.03 & 6.30 & - & - & - \\
\hline
TCDCN \cite{zhang2016learning} & 38.23 & 7.41 & 4.80 & 8.60 & 5.54 & 17.06 & 19 & 19 \\
\hline
RTSM & 6.40 & 10.93 &6.02 &16.52 & 8.06 & 8.46 & 42, 119* &  67, 190* \\
\hline
\end{tabular}}
\end{center}
\end{table*}

As addressed in Sec.\ref{sec:review}, we design the c-TSM (coarse TSM) for comparison purpose. The r-TSM and BRM are defined on faces of $100^2$ pixels, and the c-TSM is defined on faces of $50^2$ pixels. The part size is chosen 6$\times$6 for c-TSM and 10$\times$10 for r-TSM as it is 20$\times$20 for the TSM \cite{cvpr12_zhu}. The settings for extracting the HOG features are empirically optimized for different scales. The training set for the RTSM and c-TSM is composed of 1126 faces selected from the MPIE, 627 faces from the LFPW \cite{CVPR11_Belhumeur} and 1712 faces from the ALFW \cite{tugraz:icg:lrs:koestinger11b}. Note that the proposed FHM enables face detection and landmark localization to be trained independently, as the Faster R-CNN requires a huge training set, but the RTSM can be trained using thousands of training samples.

To emphasize the performance for handling in-the-wild conditions, we choose the AFW \cite{cvpr12_zhu} and the PASCAL Face \cite{Mathias2014Eccv} for evaluating face detection, and the AFW and 300W \cite{300w} for evaluating landmark localization. The PASCAL Face is used only for studying face detection as its samples are not landmark annotated. Although the 300W dataset is generally accepted as a good benchmark for assessing landmark localization, it does not contain samples with poses large than $45^o$ in yaw, which restrains its effectiveness for evaluating the landmark localization on poses beyond that range. 
The AFW offers 468 in-the-wild faces with profile-to-profile poses, various illumination conditions and facial expressions. We split the AFW into $\le45^\circ$ and $>45^\circ$ subsets to highlight the performance on extreme poses.

\subsection{Experimental Results}
\label{sec_exp1r}
The face detection performance of the Faster R-CNN, in terms of the precision-recall rates, on the AFW and the PASCAL Face are shown in Figures~\ref{ap_afw} and \ref{ap_pascal}, together with the performances of other approaches, including the c-TSM only, the c-TSM followed by the RTSM (c-TSM+RTSM), and three state-of-the-art methods, the TSM \cite{cvpr12_zhu}, the DPM and HeadHunter \cite{Mathias2014Eccv}. We follow the evaluation protocol proposed in the development of the DPM and HeadHunter detectors \cite{Mathias2014Eccv} when making Figures~\ref{ap_afw} and \ref{ap_pascal} for showing performances in precision and recall. The performance comparison on the PASCAL Face is shown in Figure~\ref{ap_pascal}. Both figures show the following observations.
\begin{enumerate}
	\item The Faster R-CNN face detector performs much better than the c-TSM+RTSM, and it is highly competitive to the state of the art with AP 97.07\% on the AFW and 94.45\% on the PASCAL Face. It clearly outperforms the DPM and HeadHunter \cite{Mathias2014Eccv} on the PASCAL Face.
	\item The c-TSM can be a coarse face detector, but it must be followed by the RTSM to remove false positives. 
	However, due to its low complexity, the c-TSM often generates many false positives, which extend the RTSM processing time.
\end{enumerate}
When tested on the AFW, the Faster R-CNN takes average 65 ms per image, the c-TSM takes 1.64 sec. and the c-TSM+RTSM takes 2.38 sec. When tested on the MPIE, in which each image only contains a face, the Faster R-CNN takes 55 ms per image, the c-TSM takes 0.92 sec. and the c-TSM+RTSM takes 1.41 sec. However, this comparison does not seem to be a fair one as the c-TSM+RTSM combines both face detection and landmark localization. For a fair comparison, we provide the faces detected by the Faster R-CNN to the state-of-the-art landmark detectors, and show their performances in Table~\ref{tab_LM}.


When calculating the landmark localization error, we adopt a common metric \cite{Belhumeur11, Shaoqing14} that normalizes the location error to the (horizontal) distance between the eyes for poses with yaw angle $<45^o$, or to the (vertical) distance between the eye and the mouth for poses beyond that range. This normalized location error is averaged over all landmarks and images in a dataset, and represented in terms of percentage.

The codes of the state-of-the-art approaches in Table~\ref{tab_LM} are mostly released by the authors, except the CDSM \cite{cdsm} and CMC-CNN \cite{hou2015facial}, which we do not have the codes and obtain its performance directly from their papers. The CDSM reports performance on AFW only, and the CMC-CNN reports on 300W only. For the five approaches that we have codes, the errors and the runtime speeds in the table are based on our tests on the same platform. The DRMF codes cease to run when presented a face with yaw angle $>45^o$, so we leave the corresponding grids empty. Since the 300W dataset can be split into a Common subset and a Challenging subset \cite{Shaoqing14}, we report the performances of the two subsets and of the overall full set. The Runtime/Face is the average time needed for landmark localization on each face. As some codes are in Matlab and some are in C/C++, we tag a star "*" to those in Matlab, and we have both types of codes for the RTSM and TSM. Note that the TSM and the CDSM combine face detection and landmark localization in the model, we put their runtime in a parenthesis $(\cdot)$. The performances shown in Table~\ref{tab_LM} can be summarized as follows.
\begin{figure}[t]
\centering
\includegraphics[width=8.6cm]{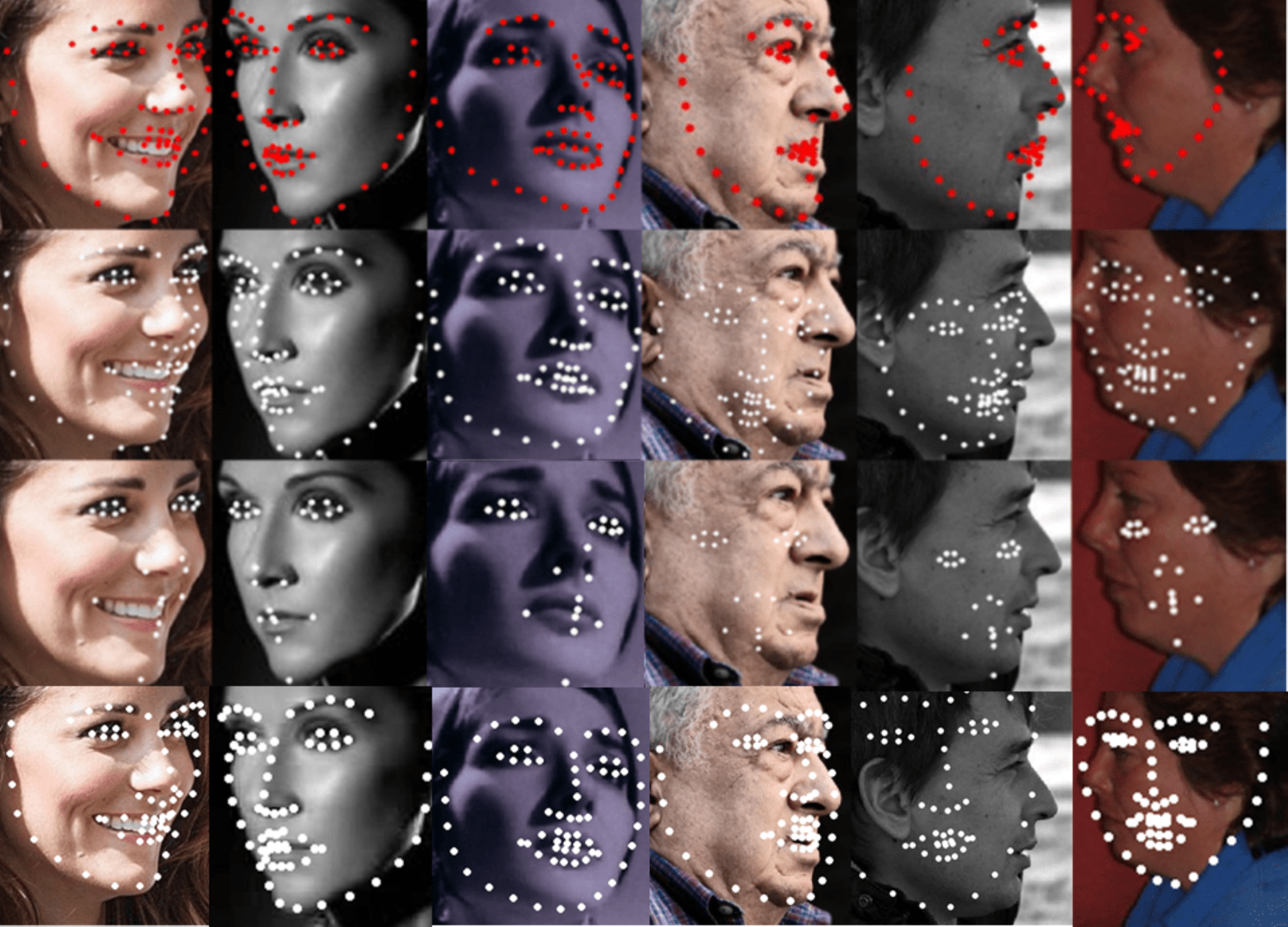}  
\caption{Comparison of different landmark localization approaches. The first row is obtained by FHM, the second by RLBF \cite{Shaoqing14}, the third by SDM \cite{Xuehan13} and the bottom by TCDCN \cite{zhang2016learning}. RLBF, SDM and TCDCN fail to handle profile or nearly profile faces, but FHM works well for all poses.}
\label{result}
\end{figure}
\begin{enumerate}
	\item The proposed RTSM (or FHM without the face detector) performs the best for poses $>45^o$, as the SDM \cite{Xuehan13}, DRMF \cite{CVPR13_Asthana}, TCDCN \cite{zhang2016learning} and RLBF \cite{Shaoqing14} cannot handle this pose range. This capacity makes the FHM one of the most appropriate landmark localization approaches for handling cross-pose recognition, which must consider extreme poses. Figure~\ref{result} shows samples with landmarks located by the best four algorithms, the FHM, RLBF, SDM and TCDCN in our test.
	\item Although the RTSM is slower than the RLBF, SDM and TCDCN, the embedded Faster R-CNN makes the FHM a favorable choice among the three when considering a total solution for face detection and landmark localization. 
	Note that the original codes for the RLBF and SDM are accompanied with the Viola-Jones face detector, which cannot detect profile or nearly profile faces. We replaced it by the Faster R-CNN face detector for a fair comparison. 
	Therefore, the numbers in Table~\ref{tab_LM} were obtained using the same set of detected faces, except for TSM and CDSM which integrate face detection and landmark in a unified model.
	\item When considering the poses $\le45^o$, the RLBF, SDM and TCDCN are among the most competitive algorithms, in both accuracy and processing time. However, for dealing with recognition for poses $>45^o$, it demands efforts to modify 
	the model templates. These approaches optimize the locations of a fixed number of landmarks as defined by the original template. It can be difficult to consider multiple templates with different numbers of landmarks for different pose ranges.
\end{enumerate}
\begin{figure}[t]
\centering
\includegraphics[width=8.6cm]{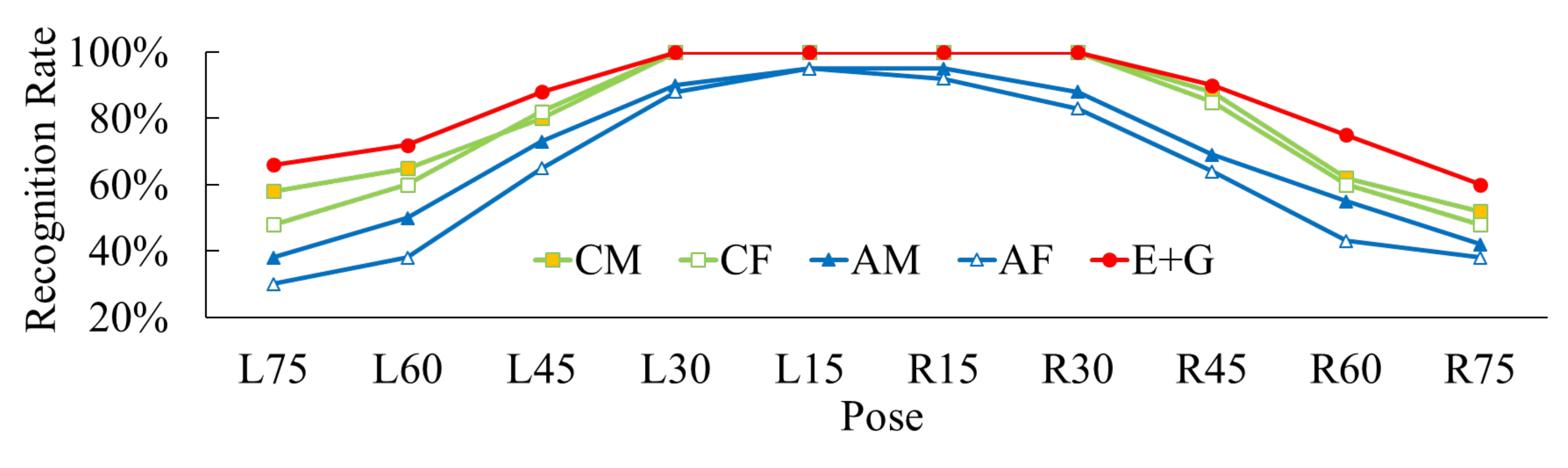}  
\caption{Comparison of cases with specific and E+G (ethnicity and gender) oriented reference model on MPIE. Four different references include Caucasian male (CM) and female (CM), Asian male (AM) and female (AF).}
\label{fig_ref_model}
\end{figure}
%
%

\section{Performance Evaluation for Cross-Pose Recognition}
\label{sec_exp2}
In this section, we present the evaluation of the component-based SRC and the CNN-based solutions for face recognition. As mentioned above, only holistic faces are studied in the CNN solution as the datasets used for training the deep network are too large and too complex to have their faces segmented into components.
\subsection{Component-based SRC Solution}
\label{sec_exp2src}
%
The experiments were carried out on the PIE database (68 subjects), the Session 1 of the MPIE database (249 subjects), and the LFW database \cite{LFW}, and the reference models were chosen from the FRGC database in the way described in Sec.\ref{sec_3d_recons}. Each subject had one single frontal face in the gallery and the rest of the poses were all in the query set. All frontal faces were aligned and normalized in size to the eyes so that the distance between the eyes was kept in 60 pixels, other poses of the same subject were normalized so that the distance between the eyes and chin is kept the same as of the frontal. The pose range in PIE covered up to $90^\circ$ in yaw, and up to $75^\circ$ on MPIE. Experimental results were separately reported for the same and different illumination conditions.
%
%
The experiments were designed to study the following issues:
\begin{enumerate}
\item Effects caused by reference models of different ethnicity and gender. 
%
\item Comparison with the holistic counterpart of the proposed method, in which the reconstruction is carried out for the whole face. This comparison should reveal the advantages of the component-based approach over the holistic approach.
\item Comparison with other state-of-the-art 3D-based approaches. As 2D-based approaches for cross-pose recognition adopt different setups (for example, the requirement for a multi-pose training set), only 3D-based approaches are considered in this comparison.
\end{enumerate}
\begin{figure}[t]
\centering
\includegraphics[width=8.6cm]{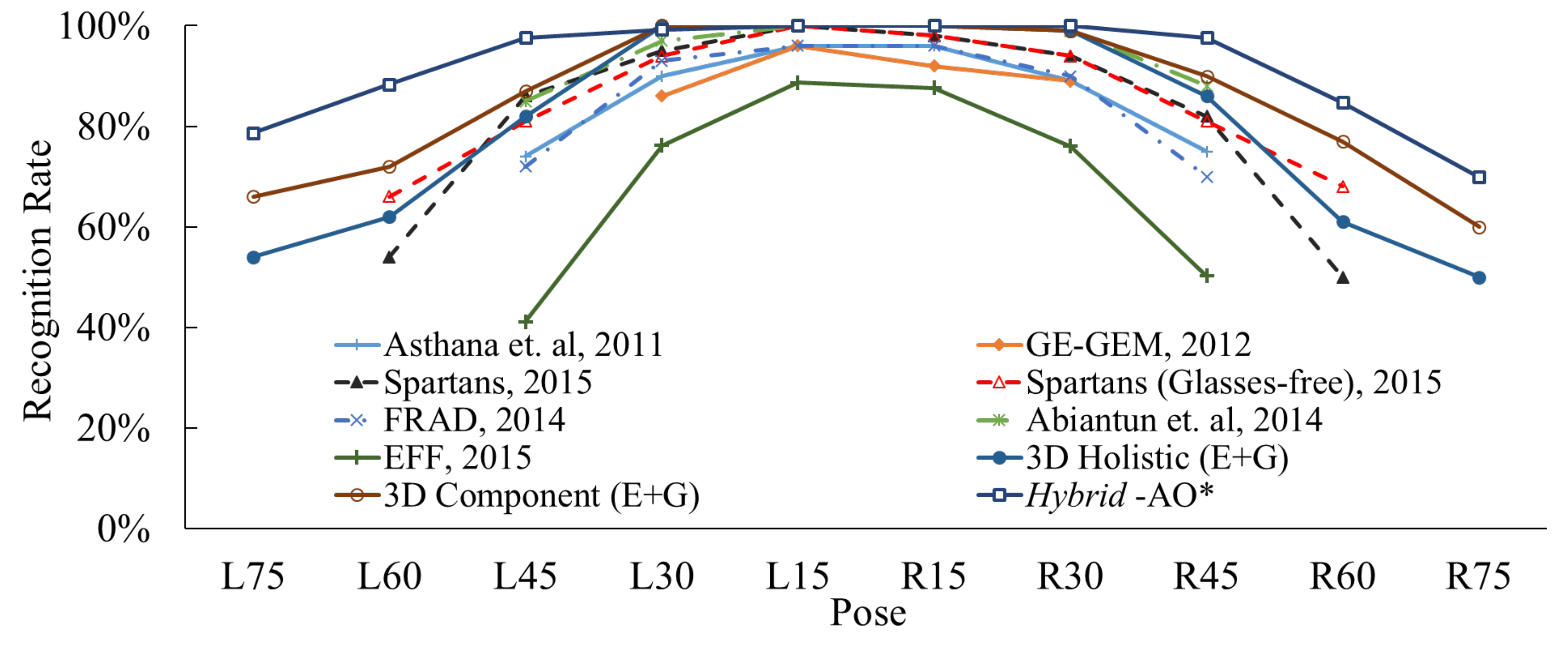}
\caption{Comparison of the proposed 3D component reconstruction, 3D holistic reconstruction, CNN-based Hybrid-AO$^*$, and contemporary 3D approaches, including Asthana et al. \cite{Asthana11}, FRAD \cite{14_TIFS_Ali}, GE-GEM \cite{HeoS12}, Spartans \cite{_15_TIP_MASM} and Abiantun et al. \cite{14_TPAMI_Sparse} on the MPIE.}
\label{fig_sota}
\end{figure}
\begin{figure}[t]
\centering
\includegraphics[width=8.6cm]{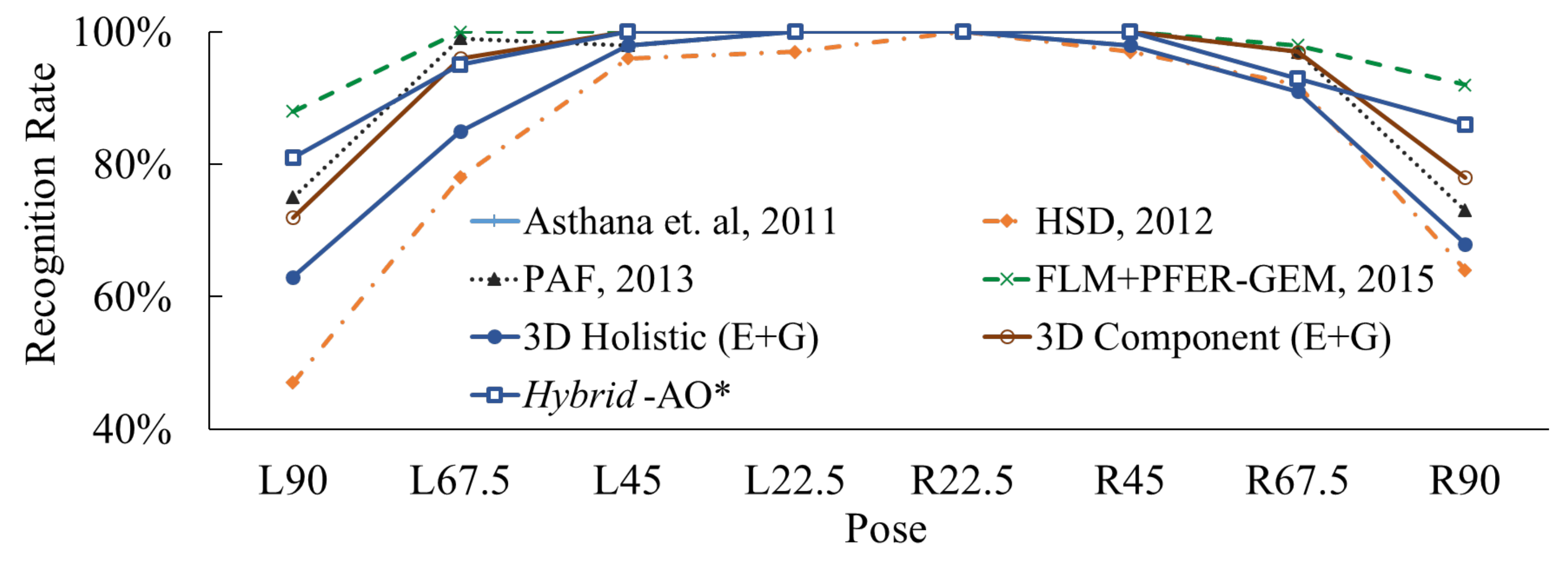}
\caption{Comparison of the proposed 3D component reconstruction, 3D holistic reconstruction, CNN-based Hybrid-AO$^*$, and contemporary 3D approaches, including HSD \cite{Zhang12}, Asthana et al.~\cite{Asthana11}, PAF~\cite{yi2013towards} and FLM+PFER-GEM \cite{_15_TIFS_clm} on the PIE database}
\label{fig_exp_pie}
\end{figure}
Figure~\ref{fig_ref_model} shows the effects caused by reference models of different gender and ethnicity. The best performance is observed when the reconstruction is based on the reference model of the same ethnicity and gender (E+G). The Caucasian male (CM) model can lead to the best performance if only one reference model is allowed. The Caucasian female (CF) comes as the close second. However, both Asian male and female models (AM and AF) perform relatively poorly. This result indicates that the effects caused by ethnicity appears much stronger than that caused by gender. This finding confirms the observation in Figure~\ref{fig_depth_samples}, which shows the depths of different reference models. These results also reflect the demographics of the MPIE, in which the largest subset is made of CMs, followed by CFs, then AMs, and least of AFs.

The proposed component-based approach outperformed its holistic counterpart in both computation cost and in accuracy. The holistic model took 2.5 minutes for one single face reconstruction, 
but the component-based model took only 18 secs. 
The comparisons of the holistic and component-based approaches with the state-of-the-art methods are shown in Figure~\ref{fig_sota} and Figure~\ref{fig_exp_pie}, on MPIE and PIE, respectively. With reference models of the same ethnicity and gender (E+G), both perform better than the contemporary methods. Quite a few 3D approaches that have been evaluated on the PIE pose subset show a significant drop in the recognition rate for yaw angle larger than 67.5$^\circ$, as shown in Figure~\ref{fig_exp_pie}. This big drop in accuracy is also observed with the 3D holistic (E+G) models. However, the 3D component (E+G) maintains its performance at 67.5$^\circ$, and it outperforms most of the contemporary methods.

Among the selected contemporary methods, the EFF (Effective Face Frontalization) \cite{hassner2015effective} is a novel approach for face frontalization. We obtained the codes released by the authors and ran experiments with the SRC-based recognition. The performance, as shown in Figure~\ref{fig_sota}, appears far from satisfactory. The {\it Hybrid}-AO$^*$ in both Figures~\ref{fig_sota} and \ref{fig_exp_pie} demonstrates the exceptional performance of our CNN solution. The details of our CNN solution are presented next.

%

%
\subsection{CNN-based Solution}
\label{sec_exp2cnn}
%
As presented in Sec.\ref{recog2}, the revised VGG network is validated with a competitive identification rate $83.51\%$ on the CASIA-WebFace. We exploit this trained network to study the performances of the following setups with the AO (Activation-at-Output) and FCF (Fully-Connected Feature) identification.
\begin{enumerate}
	\item Training on the 3D {\it Reconstructed} Models: The training set was composed of the synthesized 2D faces of all poses, generated by the 3D reconstructed facial models of the $N_g$ subjects in the gallery. The network was configured to produce $N_g$ outputs. Both the {\it Reconstructed}-AO and {\it Reconstructed}-FCF were evaluated.
	\item Training on the {\it Reference} Dataset: The network was trained on the CASIA-WebFace dataset, and exploited as a feature extractor. The 512D feature extracted from the first fully connected layer was considered a legitimate representation of a face.
	Only the {\it Reference}-FCF was tested with SVM classification in this setup.
	\item Training on the {\it Hybrid} Dataset: 
	The training set was composed of the synthesized 2D faces and the CASIA-WebFace dataset. For {\it Hybrid}-AO, the network was configured to produce $N_g+8984$ outputs. This configuration is close to an open-set identification, in which a query face can be misclassified into any of the 8984 subjects, who is not in the gallery. We compared this case with {\it Hybrid}-AO$^*$, the closed-set identification in which we disconnected the 8984 connections of the last fully connected layers to the output layer, and kept only the connections to the $N_g$ outputs. We also tested the performance of the {\it Hybrid}-FCF.
\end{enumerate}
\begin{figure}[t]
\centering
\includegraphics[width=8.6cm]{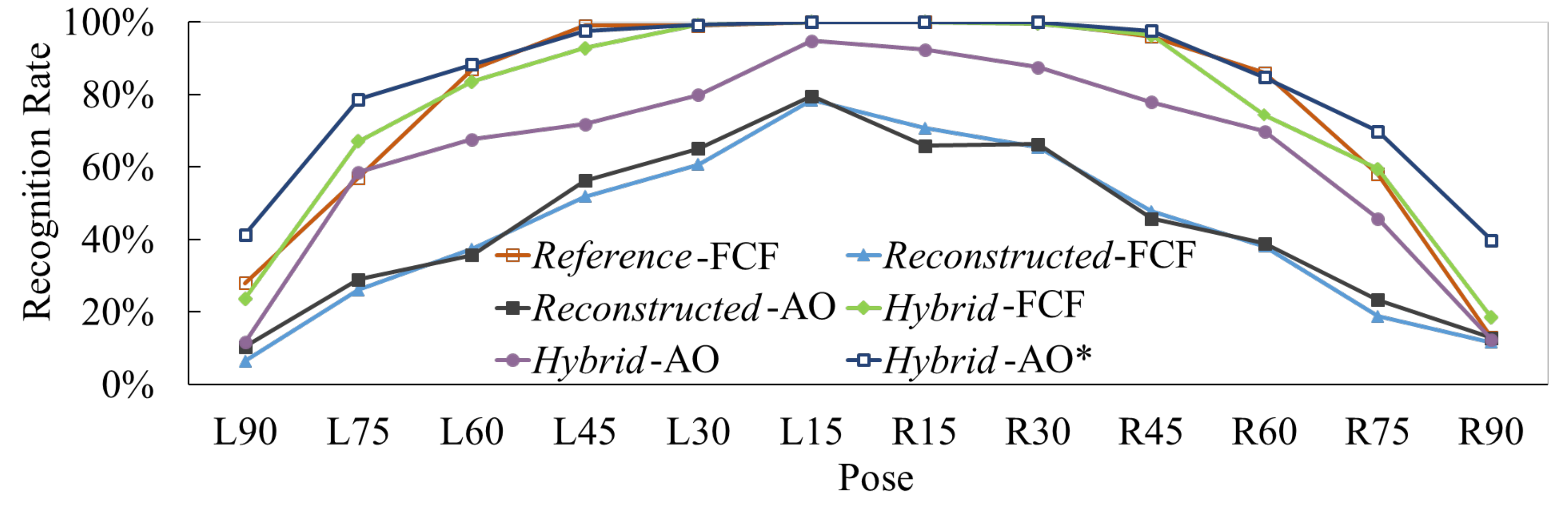}  
\caption{VGG nets with 6 different setups tested on MPIE: {\it Reconstructed}$-$ training on 2D synthesized faces generated by the 3D holistic reconstruction; {\it Reference}$-$ training on CASIA-WebFace dataset; {\it Hybrid}$-$ training on 2D synthesized faces and CASIA-WebFace. AO refers to the Activation-at-Output and FCF refers to classification using the Fully-Connected Feature.}
\label{mpie_com}
\end{figure}
%

For the tests on MPIE, the training set for the {\it Reconstructed} case is made of 158,613 synthesized faces, which were the multi-view 2D projections of the 3D reconstructed faces for the overall $N_g=$249 subjects. The performances for the above setups are shown in Figure~\ref{mpie_com}, and the observations and inferences can be summarized as follows.
\begin{enumerate}
	\item Both {\it Reconstructed}-AO and {\it Reconstructed}-FCF give the worse performances. The {\it Reference}-FCF outperforms many, especially for yaw angle $\le 60^o$. These results show that the CNN solution can be much more effective when learning from {\it in-the-wild} data, such as the CASIA-WebFace. As the samples in the CASIA-WebFace are mostly $\le 60^o$ in yaw, the identification rate of the {\it Reference}-FCF drops sharply for yaw angle $>60^o$ due to the lack of data in that pose range.
	\item The {\it Hybrid} cases generally perform well, except for the {\it Hybrid}-AO. The {\it Hybrid}-AO performs an open-set identification, in which a query image can be mistaken as one of the 8984 extra subjects in the {\it expanded} gallery. It performs poorly, even for the poses with slight rotations. The performance can be greatly improved if the network is switched to the closed-set identification, {\it Hybrid}-AO$^*$, by disconnecting the connections to the extra subjects.
	\item Although the {\it Hybrid}-AO$^*$ outperforms all, the {\it Hybrid}-FCF also performs well for all poses. The {\it Hybrid}-FCF follows the open-set identification for feature extraction, but it handles classification by using the closed-set identification with $N_g$ classes to identify. However, it is outperformed by the {\it Reference}-FCF for yaw $\le 60^o$. This shows that the features extracted from synthesized data can degrade the CNN-based performance.
\end{enumerate}

As verified in the above experiments on MPIE, the {\it Hybrid}-AO$^*$ gives the best performance. The performance comparisons with other state-of-the-art approaches are shown in Figure~\ref{fig_sota} and Figure~\ref{fig_exp_pie} for MPIE and PIE, respectively. Both figures show that the {\it Hybrid}-AO$^*$ performs better than the 3D component (E+G) approach. The {\it Hybrid}-AO$^*$ performs the best on MPIE, and the second best on PIE when considering the accuracy at $90^o$.

\begin{figure}[t]
\centering
\includegraphics[width=8.6cm]{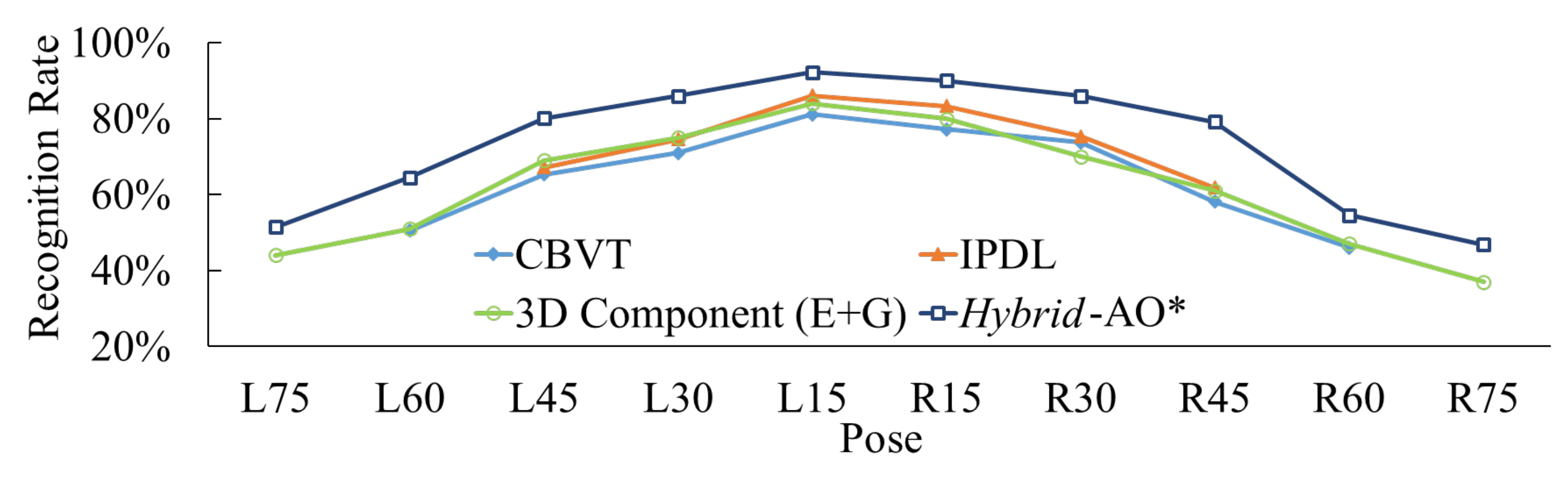}  
\caption{Performance comparison with contemporary approaches CBVT \cite{LiSG12} and IPDL \cite{zhu2013deep} for cross-pose and cross-illumination recognition on MPIE.}
\label{mpie_com_illu}
\end{figure}

The performances in Figure~\ref{fig_sota} were measured under a common scenario that a frontal facial image of each subject with uniform illumination was registered to the gallery, and the rest of poses at the same illumination were taken as query images. To evaluate the performance for handling cross-pose and cross-illumination recognition, we ran an experiment with query images covering all poses and all illumination conditions in the MPIE. This is a rarely attempted scenario and we only found it in a couple contemporary approaches, namely the Coupled Bias–Variance Tradeoff (CBVT) \cite{LiSG12} and the Identity-Preserving Deep Learning (IPDL) \cite{zhu2013deep}. The comparison of these approaches with the {\it Hybrid}-AO$^*$ and the 3D component (E+G) is shown in Figure~\ref{mpie_com_illu}. The {\it Hybrid}-AO$^*$ performs the best, with a clear gap from the rest, and the 3D component (E+G) performs similarly to CBVT and IPDL.
\begin{table}[t]
\begin{center}
\footnotesize
\caption{Face verification rates (\%) on LFW. $^*$ without E+G features added on. Selected methods are the CNN Ensemble with SAE (CESAE) \cite{Changxing2015}, the DeepFace \cite{taigman2014deepface}, the DeepID2 \cite{nips2014_Sun} and the joint Bayesian metric learning (JBML) \cite{JCChen2016}. $(\cdot)$ denotes for ensembles, 3 networks for DeepFace and 25 networks for DeepID2. For reference, human attains 97.53\% accuracy \cite{taigman2014deepface}}
\label{veri_acc}
\begin{tabular}{|c|c|c|c|c|c|}
\hline
 CESAE  & DeepFace  & DeepID2 & JBML  & $Hybrid$-FCF \\
\hline
98.43 & 95.92 (97.35) & 95.43 (99.15)  & 97.15 & 98.06, 94.52$^*$\\
\hline  
\end{tabular}
\end{center}
\label{tab1}
\end{table}

To better understand the capacity of our solutions, we have extended the experiments to face verification in which we have to determine whether a pair of faces are of the same person. This scenario is difficult to solve by using the SRC-based approach that searches for the most likely subject from a gallery set for a given query. However, this is an appropriate scenario for the feature extraction network {\it Hybrid}-FCF to handle. The database used for this study is the LFW \cite{LFW}, which contains 13,233 images of 5,749 subjects collected in the wild. 
The images in the LFW are organized into two views. View 1 is for model selection and parameter tuning, while View 2 is for performance evaluation. We follow the standard protocol and report the mean verification rate 
by running 10-fold cross validation on the View 2 subset. We use the {\it Hybrid}-FCF to extract the feature of each face, and compute the Mahalanobis distances of the match pairs and mismatch pairs in the View 1 subset. An SVM classifier with linear kernel is then trained to classify the Mahalanobis distances of both types of pairs. The performance of this SVM classifier is tested on View 2 with each pair represented in the Mahalanobis distance.
\begin{table}[t]
\begin{center}
\small { \hfill{} \hfill{} \caption{The processing time of each module in both proposed pipelines}
\begin{tabular}{|c|c|}
\hline
Face detection (Faster R-CNN, ZF Net) & 65ms \\
\hline
Training of Faster R-CNN/ZF Net on 393k faces & 12.2h \\
\hline
Landmark localization (RTSM) & 42ms \\
\hline
3D component reconstruction (per subject) & 17.6s \\
\hline
3D whole face reconstruction (per subject) & 151.6s \\
\hline
Feature extraction for a 3D component & 1.6ms \\
\hline
Identification by SRC (100 subjects in gallery) & 1.85s \\
\hline
CNN training on hybrid database (500k faces) & 42h\ 26m \\
\hline
Identification by CNN (100 subjects in gallery) & 0.397s \\
\hline
\end{tabular}}
\end{center}
\label{process_time}
\end{table}

As the 3D component reconstruction performs better with an ethnicity- and gender-oriented (E+G) reference model, we also incorporate the E+G characteristics into the VGG-extracted facial features. We train two VGG networks, the VGG-E for ethnicity classification and the VGG-G for gender classification, by using the following datasets:
\begin{enumerate}
	\item Ethnicity: 8,600 faces/race are selected for training from each of the following databases, CAS-PEAL (Asian) \cite{4404053}, Morph (African) \cite{ricanek2006morph} and PubFig (European) \cite{kumar2009attribute}. Another 1,200 faces/race are selected from each database for validation, and another 1,200 faces/race are selected for evaluation.
	\item Gender: All follow the partitions designed for the CelebA database \cite{liu2015faceattributes}, 162,770 faces for training, 19,867 for validation and 19,962 for evaluation.
\end{enumerate}

The VGG-E gains verification rate 95.11\% and VGG-G gains 97.27\%. When combining the E+G characteristics, the facial feature is concatenated with the ethnic feature and gender feature, extracted from the VGG-E and VGG-G nets, respectively. The concatenated feature is of 1,536 (512$\times$3) dimension, yielding 98.06\%. Table~\ref{veri_acc} shows the performances of the {\it Hybrid}-FCF with and without the E+G features added in, compared with state-of-the-art approaches (reviewed in Sec.\ref{sec_review_recog}), including the CNN Ensemble with SAE (CESAE) \cite{Changxing2015}, the DeepFace \cite{taigman2014deepface}, the DeepID2 \cite{nips2014_Sun} and the joint Bayesian metric learning (JBML) \cite{JCChen2016}. Note that $(\cdot)$ denotes for ensembles, 3 networks were combined for DeepFace and 25 networks combined for DeepID2, effectively boosting the performances. Considering the performance of single network, the {\it Hybrid}-FCF without E+G features performs comparably to others, it outperforms most with E+G features concatenated.

As the processing time is one of the central concerns in this study, we summarize the time spent in each component of the proposed pipelines in Table~\ref{process_time}. 


%

\section{Conclusion}
\label{con}
%
Few works like this one that reports two pipelines for tackling cross-pose recognition, one is 3D component-based and the other is CNN-based. 
The former requires only a few 3D models for depth references, but the latter needs a huge (public) face databases, e.g., CASIA-WebFace. 
We have demonstrated the following essentials: 1) The 3D component-based approach outperforms its holistic counterpart and is among the most effective solutions. 2) The VGG-based solution outperforms the hand-crafted 3D component-based approach, as the former can learn the in-the-wild characteristics that are hard to capture by the latter. 3) A comparison of the two approaches is provided to show the individual advantages. It offers a guideline for preferring one than the other when handling different scenarios. The preference can be decided based on the sufficiency of the training data. When the training data is insufficient, the 3D component-based one is preferred; when the data is abundant, the CNN-based is preferred. 4) 3D-based face reconstruction and CNN-based face model are better built with the characteristics of the same ethnicity and gender as of the subject. 5) The FHM is a vital part for landmark localization across extreme poses.

As the availability of large in-the-wild face databases is limited and the CASIA-WebFace is by far the most popular one used for CNN training, it can be a challenging and contributive research on the transformation of existing databases, e.g., MPIE and FRGC, to their in-the-wild expansions. We consider this one of the continuing research topics, and it has been carried out in our lab.

\bibliographystyle{IEEEtran}

%




\end{document}